\documentclass[preprint,12pt]{elsarticle}

\usepackage{amssymb}
\journal{Artificial Intelligence}

\usepackage{amsthm}
\newtheorem{definition}{Definition}
\newtheorem{lemma}{Lemma}

\newtheorem{theorem}{Theorem}

\newtheorem{example}{Example}

\usepackage{researchpack}
\usepackage{thmtools,thm-restate}
\usepackage{hyperref}
\usepackage{xspace}
\usepackage{xcolor}
\usepackage{booktabs}
\usepackage{tikz}
\usetikzlibrary{decorations.pathreplacing}
\usepackage{tabularx}

\newcommand{\ra}[1]{\renewcommand{\arraystretch}{#1}}

\newcommand{\add}[1]{\textcolor{black}{#1}}

\newcommand{\method}{\textsc{hassle}\xspace}
\newcommand{\methodmilp}{\textsc{hassle-milp}\xspace}
\newcommand{\methodsls}{\textsc{hassle-sls}\xspace}

\newcommand{\E}[2]{\ensuremath{\bbE_{#1}\left[#2\right]}}
\newcommand{\defeq}{\ensuremath{:=}\xspace}
\newcommand{\program}{\ensuremath{\calM}\xspace}
\newcommand{\dataset}{\ensuremath{S}\xspace}
\newcommand{\prob}{\ensuremath{\Pr}\xspace}
\newcommand{\distr}{\ensuremath{D}\xspace}
\newcommand{\context}{\ensuremath{\psi}\xspace}
\newcommand{\Contexts}{\ensuremath{\Psi}\xspace}
\newcommand{\errormatch}{\ensuremath{\chi}}
\newcommand{\regret}{\ensuremath{\text{\normalfont reg}}\xspace}
\newcommand{\score}{\ensuremath{\text{\normalfont score}}\xspace}
\newcommand{\neighbour}{\ensuremath{\text{\normalfont nbr}}\xspace}
\newcommand{\Optima}{\ensuremath{\mathrm{opt}}\xspace}
\newcommand{\wcr}{\ensuremath{r_{\mathrm{max}}}\xspace}
\newcommand{\minprob}{\ensuremath{\eta}\xspace}
\newcommand{\VC}{\ensuremath{\mathrm{VC}}\xspace}

\newcommand{\mc}{\ensuremath{\mathrm{MC}}}
\begin{document}
\begin{frontmatter}



\title{Learning MAX-SAT from Contextual Examples for Combinatorial Optimisation}

\author[kuleuven]{Mohit Kumar}
\author[kuleuven]{Samuel Kolb}
\author[unitn]{Stefano Teso}
\author[kuleuven]{Luc De Raedt}

\address[kuleuven]{KU Leuven, Belgium, firstname.lastname@cs.kuleuven.be}
\address[unitn]{University of Trento, Italy, firstname.lastname@unitn.it}

\begin{abstract}
Combinatorial optimization problems are ubiquitous in artificial intelligence. Designing the underlying models, however,  requires substantial expertise, which is  a limiting factor in practice. The models typically consist of hard and soft constraints, or combine hard constraints with an objective function. We introduce a novel setting for learning combinatorial optimisation problems from {\em contextual} examples. These positive and negative examples show -- in a particular context -- whether the solutions are good enough or not.  We develop our framework using the MAX-SAT formalism as it is simple yet powerful setting having these features.  We study the learnability of MAX-SAT models.  Our theoretical results show that high-quality MAX-SAT models can be learned from contextual examples in the realizable and agnostic settings, as long as the data satisfies an intuitive ``representativeness'' condition.  We also contribute two implementations based on our theoretical results:  one leverages ideas from syntax-guided synthesis while the other makes use of stochastic local search techniques.  The two implementations are evaluated by recovering synthetic and benchmark models from contextual examples. The experimental results support our theoretical analysis, showing that MAX-SAT models can be learned from contextual examples. Among the two implementations, the stochastic local search learner scales much better than the syntax-guided implementation while providing comparable or better models.
\end{abstract}


% \begin{highlights}
% \end{highlights}


\end{frontmatter}


% \stefano{TODO: add running example}

% \stefano{TODO: expand discussion of problem statement}

% \stefano{TODO: look at different ratios of hard vs soft constraints?  like 0-100 and 80-10}

% \stefano{TODO: does increasing \# of negative examples improve infeasibility?  is there a trade-off between accuracy and infeasibility?}

\section{Introduction}
\label{sec:introduction}

Combinatorial optimisation is an effective and popular class of techniques for solving real-life problems like scheduling~\cite{Demirovic2019}, routing~\cite{mills2000guided}, and planning~\cite{robinson2010partial}. However, encoding the underlying models often proves to be time-consuming and complicated, as it requires substantial domain and modeling expertise.  Therefore, the question arises as to whether such models can be learned from data. This question is studied in constraint learning~\cite{bessiere2016new,de2018learning}, where several algorithms have been developed that automatically acquire theories or mathematical models from examples of past working (positive) and non-working (negative) solutions or analogous forms of supervision.

Combinatorial optimisation models have two components:
a set of \emph{hard constraints} $\phi$ defining the feasible region, and
an \emph{objective function} $f$ that measures the quality of candidate solutions, sometimes defined as a set of soft constraints.
The task of the solver is then to complete a (potentially empty) partial assignment $x$ into a complete assignment $xy$ that is both feasible and optimal, i.e., $xy \models \phi$ and $y \in \argmax_{x y \models \phi} f(xy)$.  We use the term \emph{context} to refer to both partial assignments $x$ and to more general temporary constraints that restrict the outcome of optimisation.

Current learning approaches suffer from two limitations.  First, to the best of our knowledge, they do not learn from contextual examples.  By doing so, they ignore the fact that the optima can be affected drastically by the context.  For instance, $\add{\argmax}_{x : x y \models \phi} f(x y)$, where the context fixes $y$, can be very different from $\add{\argmax}_{y : x y \models \phi} f(x y)$, where the context fixes $x$.  Furthermore, this is also a less realistic setting  in practice, as examples of good and bad solutions will always be relative to a context.  The reader may notice a resemblance with structured output prediction (e.g.~\cite{tsochantaridis2005large}), where one learns a function $f$ that computes a structured output $y = \argmax_y f(x,y)$ for a given input $x$.  The difference is that in structured prediction the choice of input and output variables is fixed, while in optimisation it is not.

Second, existing approaches do not  {\em jointly} learn the hard constraints and the objective function: they either learn one or the other, or else learn them sequentially or independently.  But this may break down in applications like personnel rostering.  Here, past schedules are often stored in a data set, but the reasons why a schedule was found to be unacceptable is usually not tracked.  In cases like this, a negative example may be either infeasible (because of the hard constraints) or sub-optimal (because of the objective function).  This induces a credit-assignment problem that can only be solved by learning the constraints and objective function jointly.  See the related work section for a more in-depth discussion.

The key contribution of this paper is that we develop a more realistic setting for learning combinatorial optimisation models {\em from contextual examples} that does not suffer from these limitations.   Furthermore, we provide foundational results within this setting for one of the simplest but most fundamental models for combinatorial optimisation, maximum satisfiability (MAX-SAT for short).  Our theoretical results show that MAX-SAT models can be probably approximately correctly (PAC) and agnostically learned from contextual data using empirical risk minimization (ERM) as long as the contextual examples are ``representative enough'', and that if enough data is available the acquired model is guaranteed to output high-quality feasible solutions.

Motivated by this, we introduce two implementations of ERM for MAX-SAT learning, \methodmilp and \methodsls.  \methodmilp relies on ideas from syntax-guided synthesis~\cite{alur2018search}, in that the learning task is encoded as an optimization problem -- namely, a mixed-integer linear programming (MILP) problem -- and solved using an efficient solver.  \methodmilp acquires a MAX-SAT model that is guaranteed to fit the examples (almost) exactly, if one exists.  This accuracy, however, comes at the expense of run-time.  The second implementation, \methodsls, uses stochastic local search (SLS) to look for a high-quality MAX-SAT model, and in addition integrates a heuristic to prune the neighborhood of the current candidate model and focus on the most promising neighbors.  \methodsls is not guaranteed to return an (almost) optimal model, but it offers enhanced efficiency.  Our experiments show that, on the one hand, \methodmilp successfully recovers both synthetic and benchmark MAX-SAT models from contextual examples, and on the other, that \methodsls matches the model quality of \methodmilp in a fraction of the time and scales to learning problems beyond the reach of the more exact implementation.

A preliminary version of this work appeared as a conference paper~\cite{kumar2019learning}.  The present work contributes the following major improvements:
\begin{itemize}

    \item The theoretical results in the conference paper hold for the realizable, noiseless case only, and only guarantee that, given enough examples, the learned model will perform well at \add{solving} time in the presence of the empty context.  Here, we show that, under mild assumptions, these results apply to the agnostic and noisy cases.  In addition, we show that the learned model will perform well in \emph{arbitrary} contexts at \add{solving}-time and introduce a tighter regret bound, see Theorem~\ref{thm:RegretBound}.

    \item  The only implementation available in the conference paper is \methodmilp.  Here we introduce a new implementation based on stochastic local search~\cite{hoos2004stochastic}, named \methodsls.  In particular we studied four different versions of \methodsls, based on some of the most widely used SLS techniques~\cite{Hoos2015}:  WalkSAT, Novelty, Novelty$^+$ and Adaptive Novelty$^+$.  Compared to the original implementation, \methodsls scales to larger learning tasks while acquiring models of comparable or better quality in practice.  The implementation is non-trivial as evaluating the score of each neighbour requires solving a MAX-SAT \add{model}.  To keep the run-time low, we designed techniques to restrict the neighbourhood to those models which show some promise of being better than the current candidate.

\end{itemize}

This paper is structured as follows: Section~\ref{sec:pre} provides notations and definitions for various terms used throughout the text and Section~\ref{sec:problem} provides a formal definition of MAX-SAT learning.  In Section~\ref{sec:learnability}, we first prove PAC learnability of MAX-SAT learning and that this provides guarantees on the quality of the \add{assignment} output by the learned model (relative to those output by the ground-truth model).  In Section~\ref{sec:hassle}, we present the two implementations and evaluate them in Section~\ref{sec:exp}.  The related work in discussed in Section~\ref{sec:related} and the paper concludes in Section~\ref{sec:conclusion}.  For ease of exposition, all the proofs are deferred to the Appendices. A summary of all the notations is provided in Table~\ref{tab:notations}.

\section{Preliminaries}
\label{sec:pre}

\subsection{Maximum Satisfiability}

Let $\vX = \{X_1, \ldots, X_n\}$ be a set of Boolean variables and $\Phi = \{\phi_1, \ldots, \phi_m\}$ a class of Boolean formulas of interest on $\vX$, e.g., the set of conjunctions or disjunctions of up to $k$ literals (i.e., variables or their negations).  An assignment $\vx = (x_1, \ldots, x_n)$ fixes the value of each $X_i$ to $x_i$.  In practical implementations, we will use $1$ and $0$ to encode true and false, respectively.

A \emph{partial maximum satisfiability} (abbreviated MAX-SAT) \add{model} \program is a collection of hard and soft constraints taken from $\Phi$~\cite{miyazaki1996database,li2009maxsat}:  the hard constraints define a feasible region while the soft ones define a preference relation over feasible \add{assignments}.  The hard constraints are encoded by a vector $\vc \in \{0, 1\}^m$ such that $c_j = 1$ if $\phi_j$ appears as a hard constraint in \program and zero otherwise.  The conjunction of all hard constraints is indicated as:
$$
    \textstyle
    \phi(\vc) \; \defeq \; ((c_1 = 1) \Rightarrow \phi_1) \land \ldots \land ((c_m = 1) \Rightarrow \phi_m)
$$
The soft constraints are encoded as a weight vector $\vw \in [-1, 1]^m$, where $w_j \ne 0$ if $\phi_j$ appears as a soft constraint with weight $w_j$ and $w_j = 0$ otherwise.  Hence, each constraint $\phi_j \in \Phi$ can be hard ($c_j = 1$ and $w_j$ arbitrary), soft ($c_j = 0$ and $w_j\ne 0$), or irrelevant ($c_j = w_j = 0$).  All MAX-SAT \add{models} can be written in this form by normalizing their weights.

The value $f_{\vw}(\vx)$ of an assignment $\vx$ is the total weight of the constraints that it satisfies, that is\footnote{The weights of hard constraints can be safely ignored, as they offset the value of all feasible \add{assignments} by the same amount.}:
\begin{equation}
    f_{\vw}(\vx) \defeq \sum_{j = 1}^m w_j \Ind{\vx \models \phi_j}
    \label{eq:objective}
\end{equation}
where the indicator function $\Ind{cond}$ evaluates to 1 if $cond$ holds and to $0$ otherwise.\footnote{\add{Notice that the range of the weights $\vw$ is immaterial: it is always possible to rescale the weights to any range $[a, b]$ without altering the ranking of feasible assignments (namely, by offsetting and multiplying the value of all assignments by appropriate amounts).  As a matter of fact, our algorithms learn weights in the range $[0, 1]$.  We keep negative weights here to stress that they can be used to encode negative preferences.}}   \add{Solving} a MAX-SAT model amounts to finding a feasible \add{assignment} that has optimal value, that is:
\begin{align}
    \add{\argmax}_{\vx}
        & \ f_{\vw}(\vx)
        \label{eq:inference}
    \\
    \text{s.t.}
        & \ \vx \models \phi(\vc)
        \nonumber
\end{align}
or $\add{\argmax}_{\vx \models \phi(\vc)} f_{\vw}(\vx)$ for short.  Here $\vx \models \phi$ indicates that $\vx$ satisfies $\phi$.

\begin{example}
Consider a nurse rostering problem where $N$ is the set of nurses and \add{$T$} is the set of shifts in a day.  Let $X_{n,\add{t}}$ be a Boolean variable that takes value 1 if nurse $n \in N$ is scheduled for shift \add{$t \in T$} and 0 otherwise. Consider a set of soft constraints which ensures that \add{the total number of working shifts} is minimal, and a set of hard constraints ensuring that each shift has a nurse. These can be encoded as follows:
\begin{align*}
     \text{Soft Constraints: } & \forall n \in N, \add{t \in T} :
        \neg X_{n,\add{t}} \; \text{with weight $w_{n,\add{t}} = 1$}\\
    \text{Hard Constraints: } & \forall \add{t \in T} :
        \bigvee_{n \in N} X_{n,\add{t}} \; 
\end{align*}
Then the possible assignments to $\vX = \{X_{n,\add{t}} : n \in N, \add{t \in T}\}$ represent the set of schedules for all nurses and shifts in a single day.  Clearly, an optimal assignment $\vx$ satisfies the hard constraints and maximizes the total weight of the satisfied soft constraints $f(\vx) = \sum_{n,\add{t}} 1 \cdot \Ind{\vx \models \neg x_{n,\add{t}}}$.
\end{example}

Although solving MAX-SAT is NP-hard in general~\cite{li2009maxsat}, practical solvers have recently shown impressive performance on highly non-trivial instances.\footnote{See for instance: \url{https://maxsat-evaluations.github.io}}

\subsection{Contexts}

In real-life applications, decisions are influenced by temporary conditions like resource availability and other kinds of restrictions.  For instance, in personnel scheduling some employees may be unavailable because they are sick or on leave, while in routing tasks part of the network may be down due to a temporary failure.

Such restrictions are captured by the notion of \emph{context}.  As in the previous examples, a context \context fixes the value of one or more decision variables and can be viewed as a conjunction of literals.  Letting \program be a MAX-SAT model, \add{solving} in a context \context amounts to:
\begin{equation}
    %\textstyle
    \add{\argmax}_{\vx \,\models\, \phi(\vc) \,\land\, \context} \; f_{\vw}(\vx)
    \label{eq:csinference}
\end{equation}
Note that the context, as well as the set of variables it fixes (if any), is likely to change over time.

Contexts can do more than specifying partial \add{assignment}.  In rostering applications, for instance, an employee may be unable to work more than four hours a day due to various contingencies (e.g., pregnancy), while in routing some paths may be shut down due to accidents.  In the most general sense, a context \context specifies an arbitrary Boolean formula that does restrict -- but does not necessarily fix -- some decision variables.  For example, the fact that one out of several radiologists is absent is best represented as a disjunction.

\begin{example}
Consider our nurse rostering problem.  In this case, a context could be that a nurse $n_1$ was on leave on a particular day, i.e., $\context= \bigwedge_{t \in T}\neg X_{n_1,t}$ or that the hospital was closed for shift $t_2$, i.e., $\context= \bigwedge_{n \in N}\neg X_{n,t_2}$.
\end{example}

\add{In the most general terms, contexts specify temporary constraints that a particular (set of) example(s) was obtained in.  Notice that, in contrast to the constraints that define the MAX-SAT model, they are supplied together with the data rather than learned from the examples.}

Contexts are central in optimization problems as they can significantly alter the properties and quality of optima: optimal schedules may be substantially better if a particularly skilled worker is available, and much worse otherwise.  From a learning perspective, contexts change the way a model is evaluated:  a model that performs well in a context might perform poorly in other contexts.  This means that failing to take contexts into account may lead to evaluate candidate MAX-SAT models poorly and therefore to prefer under-performing models over better ones.

\section{Problem Statement}
\label{sec:problem}

In contrast to existing constraint learning methods, we consider a more realistic setting where example solutions and non-solutions are \emph{context-specific}.  Here and below, $\Contexts^*$ indicates the set of possible contexts.  More specifically, we assume each example to be generated as follows:
\begin{enumerate}

    \item A \emph{context} $\context \in \Contexts^*$ is observed;

    \item A \emph{\add{assignment}} $\vx$ that satisfies context \context (i.e., $\vx \models \context$) is chosen according to some policy, e.g., by asking a domain expert to provide either a high-quality solution or a non-solution;

    \item $\vx$ is \emph{labelled} as positive, $y = 1$, if it works well enough in practice \add{under the observed temporary restrictions, i.e., context $\psi$}, and is labelled negative $y = 0$ otherwise.

\end{enumerate}
This induces a ground-truth distribution
$
    \distr(y, \vx, \context) = \distr(\context) \distr(\vx \mid \context) \distr(y \mid \vx, \context)
$.
We are now ready to define our learning task:

\begin{definition}[MAX-SAT learning]
    Given a set of Boolean variables $\vX$, candidate constraints $\Phi$, and context-specific examples $\dataset \defeq \{(\context_k, \vx_k, y_k) \,:\, k = 1, \ldots, s\}$ sampled \add{independently and identically distributed (i.i.d.)} from an unknown distribution \distr, find a MAX-SAT model \program with parameters $\vc$, $\vw$ that can be used to obtain high-quality \add{assignments} in any context \context.
\end{definition}

The exact nature of \add{assignment} quality will be formalized in the next section using the notion of regret~\cite{cesa2006prediction}.

A few remarks are in order.  First, we do not know which negative examples are infeasible and which ones are sub-optimal.  This introduces a \emph{credit assignment} problem: if $\vx$ is a false positive, should $\vc$ be changed to make it infeasible or should $\vw$ be adapted to make it sub-optimal? For this reason, the soft and hard constraints have to be acquired simultaneously.

Second, all \add{assignments} are expected to include a description of the context in which they were obtained.  This is often the case in real-life applications:  in rostering, past schedules are annotated with the unavailable employees, while in routing, the conditions under which the congestion occurred (e.g., maintenance work) are kept track of for monitoring purposes.  Applications in which context annotations are not available could be addressed by treating the unobserved contexts as latent variables.  This extension is non-trivial and therefore left to future work. Below we provide an example of our learning task.
\begin{example}
Consider our nurse rostering problem. Let us assume that $N=\{n_1,n_2\}$ and $T=\{t_1,t_2\}$, then $\vX=\{X_{n_1,t_1},X_{n_1,t_2},X_{n_2,t_1},X_{n_2,t_2}\}$. Now consider these context specific examples, where each example consists of three elements, first is the observed context, second is an assignment to the variables in $\vX$ and finally, third is the label specifying whether the assignment is a positive or negative sample in the observed context:
$$
    \dataset=\{(\neg X_{n_1,t_2}\land \neg X_{n_2,t_2}, \{1,0,0,0\}, 1),(\neg X_{n_1,t_2}\land \neg X_{n_2,t_2}, \{1,0,1,0\}, 0)\}
$$
Both examples correspond to the context that the hospital is closed for shift $t_2$. In this context, the first example is a positive while the second one is a negative.
The aim of our work is to learn the soft and hard constraints, provided such context specific examples.
\end{example}

\begin{table}[tb]
    \centering
    \caption{Summary of the notation used in this paper.}
    \vspace{2mm}
    \begin{footnotesize}
    \begin{tabular}{ m{3cm}  m{9cm} }
        \toprule
        \textbf{Symbol} & \textbf{Meaning}
        \\
        \midrule
        $\calX \subseteq \{\top, \bot\}^n$
            & Domain of possible assignments. \\
        $\vX = \{X_1, \ldots, X_n\}$
            & Decision variables.  All Boolean. \\
        $\vx = (x_1, \ldots, x_n)$
            & Assignment to \vX. \\
        $\Phi = \{\phi_1, \ldots, \phi_m\}$
            & Set of candidate Boolean formulas on \vX. \\
        \midrule
        $\program$
            & MAX-SAT model \\
        $\vc \in \{0, 1\}^m$
            & $c_j = 1$ if $\phi_j$ is a hard constraint in \program and 0 otherwise. \\
        $\vw \in [-1, 1]^m$
            & Weight of constraint $\phi_j$.  Only meaningful if $c_j = 0$. \\
        $h: \calX \rightarrow \{0, 1\}$
            & MAX-SAT classifier corresponding to model \program. \\
        $\calH$
            & Class of candidate MAX-SAT classifiers. \\
        $\context$
            & Context. \\
        $\Contexts^*$
            & Set of all possible contexts. \\
        \midrule
        $\program^*$
            & Ground-truth MAX-SAT model. \\
        $h^*$
            & Ground-truth MAX-SAT classifier. \\
        $\dataset \defeq \{(\context_k, \vx_k, y_k)\}$
            & Training set. \\
        $y_k$
            & Ground-truth label of $\vx_k$.  If noiseless, $y_k = h^*(\vx_k)$. \\
        \bottomrule
    \end{tabular}
    \end{footnotesize}
    \label{tab:notations}
\end{table}

\section{Learnability of MAX-SAT Models}
\label{sec:learnability}

In this section, we study the learnability of MAX-SAT models from contextual examples from the perspective of statistical learning theory.\footnote{\add{As is common in this literature, we focus on statistical aspects only.  MAX-SAT learning, just like other forms of structure learning, is computationally hard.}}  Basic knowledge of statistical learning theory is assumed;  the necessary material, \add{including the definitions of Vapnik-Chervonenkis dimension and the theorems that link it to generalization outside of the training set}, can be found in Shalev-Shwartz and Ben-David~\cite{shalev2014understanding}.

Our analysis relies on a reduction of MAX-SAT learning to learning binary ``MAX-SAT classifiers'', to which results from statistical learning theory apply.  The analysis is split into three steps:
\begin{enumerate}

    \item In Section~\ref{sec:intracontext} we define MAX-SAT classifiers and show that they can be learned from context-specific examples using Empirical Risk Minimization~\cite{vapnik1992principles}.

    \item In Section~\ref{sec:clftoopt} we prove that MAX-SAT classifiers with low context-specific risk correspond to MAX-SAT models with high-quality context-specific solutions.

    \item Finally, in Section~\ref{sec:intercontext} we show that, under a suitable ``representativeness'' condition, MAX-SAT classifiers with low context-specific risk on the contexts annotated in the data enjoy low context-specific risk in any target context.

\end{enumerate}
Taken together, these results entail that MAX-SAT classifiers learned from enough context-specific examples correspond (again, under suitable assumptions) to MAX-SAT models that produce high-quality solutions in any target context.  This shows that Empirical Risk Minimization (ERM) is a valid solution strategy for MAX-SAT learning and motivates our ERM-based implementations, introduced in the next Section.

\subsection{Prerequisites}

Before proceeding, we provide a brief summary of some key results from statistical learning theory~\cite{shalev2014understanding}.
Let $\calH$ be a class of binary classifiers (aka hypotheses) $h: \calX \rightarrow \{0, 1\}$, \add{where $\calX$ is the set of assignments to decision variables in $\vX$}, and let $h^*$ be the ground-truth classifier used to label a training set $\dataset$.  If $h^* \in \calH$ we say that the setting is \emph{realizable}, otherwise we call the setting \emph{agnostic}.  Given a hypothesis $h \in \calH$, the \emph{risk} $L_{\distr}(h)$ of $h$ is the probability that it misclassifies an \add{assignment}, that is:
$$
    L_{\distr}(h) \defeq \distr(h(x) \ne y)
$$
where $\distr$ is the unobserved, ground-truth distribution, while the \emph{empirical risk} $L_{\dataset}(h)$ is the probability that it misclassifies an example in \dataset, that is:
$$
    L_{\dataset}(h) \defeq \frac{1}{s} \sum_{k=1}^s \Ind{h(x_k) \ne y_k}
$$
\add{Where $s=|S|$.} The following definitions are taken from~\cite{shalev2014understanding}:

\begin{definition}[PAC learnability]
\label{def:pac}
A hypothesis class $\calH$ is PAC learnable in the realizable setting if there exists a function $s_{\calH}:(0,1)^2 \rightarrow \bbN$ and a learning algorithm such that, for every $\epsilon, \delta \in (0,1)$ and distribution $\distr$ over $\calX \times \{0, 1\}$, running the algorithm on $s \ge s_{\calH}(\epsilon,\delta)$ i.i.d. examples sampled from $\distr$ returns a hypothesis $h$ such that:
\begin{equation}
    \prob(L_{\distr}(h) > \epsilon) \le \delta \label{eq:pacrealizable}
\end{equation}
A hypothesis class $\calH$ is PAC learnable in the agnostic setting if, under the same conditions as above, it holds that:
\begin{equation}
    \prob(L_{\distr}(h) > \min_{h' \in \calH} L_{\distr}(h') + \epsilon) \le \delta \label{eq:pacagnostic}
\end{equation}
\end{definition}

PAC learnability simply means that, given enough examples, one can identify with high probability a hypothesis whose risk is very close to the risk of the best possible hypothesis.
% Onward, we use the term ``PAC learnable'' to refer to learnability in the realizable setting, unless explicitly mentioned otherwise.

\begin{definition}[$\epsilon$-representative]
A training set \dataset is $\epsilon$-representative iff:
\[
    \forall h \in \calH, \; |L_{\dataset}(h)-L_{\distr}(h)| \le \epsilon
\]
\end{definition}

\begin{definition}[Uniform convergence]
A hypothesis class $\calH$ is said to have the uniform convergence property if there exists a function $s_{\calH}:(0,1)^2 \rightarrow \bbN$ such that for every $\epsilon, \delta \in (0,1)$ and distribution \distr over $\calX$, if \dataset is a sample of $s \ge s_{\calH}(\epsilon,\delta)$ examples drawn i.i.d. from \distr, then, with probability of at least $1-\delta$, \dataset is $\epsilon$-representative.
\end{definition}

\add{The} uniform convergence property simply says that, given enough examples, the empirical risk is very close to the true risk.

% Ideally we would like a learning algorithm that outputs a hypothesis $h \in \calH$ that minimizes the true risk given by $L_{\distr}(h)$.  However, the distribution $\distr$ is unknown and the best that we can do is to optimize the empirical risk $L_{\dataset}(h)$ instead.  An Empirical Risk Minimization (ERM) algorithm does exact this:  it outputs a hypothesis $h \in \calH$ with minimal empirical risk.
It is well known that if a hypothesis class $\calH$ is ``simple'', i.e., it has finite Vapnik-Chervonenkis (VC) dimension~\cite{shalev2014understanding}, then it has the uniform convergence property:
% and any empirical risk minimizer $h \in \argmin_h L_{\dataset}(h)$ has low true risk with high probability:

\begin{theorem}[Theorems 6.7 and 6.8 from~\cite{shalev2014understanding}]
\label{thm:sampleComplexity}
Let $\calH$ be a hypothesis class with finite VC dimension $d \defeq \add{\VC}(\calH) < \infty$.  Then $\calH$ has the uniform convergence property with sample complexity:
$$
    \textstyle
    \frac{C_1}{\epsilon^2} \left(d + \log\left(\frac{1}{\delta}\right)\right) \le s_{\calH}(\epsilon,\delta) \le \frac{C_2}{\epsilon^2} \left(d \log\left(\frac{1}{\epsilon^2}\right) + \log\left(\frac{1}{\delta}\right)\right)
$$
Here $C_1$ and $C_2$ are universal constants.
\end{theorem}

%\begin{itemize}

    % \item $\calH$ is PAC learnable with any ERM learning algorithm with sample complexity:
    % $$
    %     \textstyle
    %     \frac{C_1}{\epsilon} \left(d + \log\left(\frac{1}{\delta}\right)\right) \le s_{\calH}(\epsilon,\delta) \le \frac{C_2}{\epsilon} \left(d \log\left(\frac{1}{\epsilon}\right) + \log\left(\frac{1}{\delta}\right)\right)
    % $$

    % \item $\calH$ is agnostic PAC learnable with any ERM learning algorithm with sample complexity:
    % $$
    %     \textstyle
    %     \frac{C_1}{\epsilon^2} \left(d + \log\left(\frac{1}{\delta}\right)\right) \le s_{\calH}(\epsilon,\delta) \le \frac{C_2}{\epsilon^2} \left(d \log\left(\frac{1}{\epsilon^2}\right) + \log\left(\frac{1}{\delta}\right)\right)
    % $$
    
    % \item $\calH$ has the uniform convergence property with sample complexity:
    % $$
    %     \textstyle
    %     \frac{C_1}{\epsilon^2} \left(d + \log\left(\frac{1}{\delta}\right)\right) \le s_{\calH}(\epsilon,\delta) \le \frac{C_2}{\epsilon^2} \left(d \log\left(\frac{1}{\epsilon^2}\right) + \log\left(\frac{1}{\delta}\right)\right)
    % $$
    
    % \item Any Empirical Risk Minimization (ERM) rule is a successful PAC learner for $\calH$

%\end{itemize}

The theorem entails that searching for a hypothesis with minimal empirical risk -- a strategy called Empirical Risk Minimization (ERM) -- is guaranteed to find a hypothesis with low risk in the realizable and agnostic settings with high probability, as long as enough examples are available.  ERM achieves similar results even if the example labels are flipped (due to noise) independently at random with probability $0 \le \rho < \frac{1}{2}$, in which case the sample complexity is asymptotically identical to the one for the agnostic setting except for an extra factor $(1 - 2\rho)^{-2}$, cf.~\cite{angluin1988learning}.

\subsection{Learnability in a Fixed Context}
\label{sec:intracontext}

We start by introducing the central notion of MAX-SAT classifier.
\begin{definition}[MAX-SAT classifier]
Given a MAX-SAT model \program with parameters $\vc$ and $\vw$, the corresponding MAX-SAT classifier $h_{\vc,\vw}: \calX \times \Contexts^* \to \{0, 1\}$ is defined as:
$$
    %\textstyle
    h_{\vc,\vw}(\vx,\context) = \Ind{\vx \in \argmax_{\vx' \; \models \; \phi(\vc) \; \land \; \context} f_{\vw}(\vx')}
$$
\label{def:model}
\end{definition}
In other words, a MAX-SAT classifier $h_{\vc,\vw}(\vx,\context)$ labels \add{assignments} $\vx$ as positive if and only if they are solutions to the corresponding MAX-SAT model, that is, \emph{feasible} with respect to $\phi(\vc) \,\land\, \context$ and \emph{optimal} with respect to $f_{\vw}$.

The set of MAX-SAT classifiers will be henceforth referred to as:
$$
    \calH \defeq \{ h_{\vc,\vw} : \vc \in \{0, 1\}^m, \vw \in [-1, 1]^m \}
$$
In the following, we momentarily focus on MAX-SAT classifiers restricted to a single context $\context \in \Contexts^*$:
$$
    \calH_\context \defeq \{h(\cdot,\context) : h \in \calH\}
$$
Our first result is that, regardless of the choice of context \context, $\calH_\context$ has finite VC dimension.

\begin{theorem}
\label{thm:ci}
For every \context, it holds that $\add{\VC}(\calH_\context) \le u (2m + 1)$, where $u < 5$.
%\stefano{note that $m$ can be quite large.}
\end{theorem}

All proofs can be found in the Appendix. Now, let $L_{\dataset,\context}$ be the empirical risk of $h$ in the context \context, that is:
$$
    L_{\dataset,\context}(h) \defeq \frac{1}{s} \sum_{k=1}^s \Ind{h(\vx_k,\context) \ne y_k}
$$
Similarly, let $L_{\distr,\context}(h)$ be the true risk of $h$ w.r.t. context \context, that is:
\begin{align*}
    L_{\distr,\context}(h)
        & \defeq \E{y,\vx\mid\context}{\Ind{h(\vx,\context) \ne y}} = \sum_{\vx \models \context, \, y} \Ind{h(\vx, \context) \ne y} \distr(\vx,y\mid\context)
\end{align*}
Taken together, Theorems~\ref{thm:sampleComplexity} and \ref{thm:ci} show that $\calH_\context$ has the uniform convergence property:

\begin{restatable}{corollary}{ThmUC}
\label{thm:uc}
For any $\context \in \Contexts^*$, and $\epsilon_\context$, $\delta_\context \in [0, 1]$, there exists an integer $s_{\calH_\context}(\epsilon_\context,\delta_\context) \in \bbN$ such that, given a dataset \dataset with at least $s_{\calH_\context}(\epsilon_\context,\delta_\context)$ examples specific to \context, any hypothesis $h \in \calH_\context$ satisfies the following:
$$
    \add{\prob(L_{\distr,\context}(h) < L_{\dataset,\context}(h) + \epsilon_\context) \ge 1-\delta_\context}
$$
% In the realizable setting the above inequality reduces to:
% $$
% \prob(L_{\distr,\context}(h) > \epsilon_\context) \le \delta_\context
% $$
The sample complexity $s_{\calH_\context}(\epsilon_\context,\delta_\context)$ is given by Theorem~\ref{thm:sampleComplexity}.
\end{restatable}

In turn, this means that $\calH_\context$ is PAC learnable:  given enough examples, with high probability the true risk is bounded by the empirical risk and ERM outputs a low-risk hypothesis.

\subsection{From Classification to Optimization}
\label{sec:clftoopt}

We have just shown that ERM can be used to learn low-risk MAX-SAT \emph{classifiers} for any given context.  Next, we show that these classifiers correspond to good MAX-SAT \emph{models}, that is, models useful for combinatorial optimization.

% In the remainder of the section, we will consider the agnostic setting, we represent the best possible hypothesis by $h^*$ -- with parameters $\vc^*$, $\vw^*$: $h^* = \mathrm{argmin}_{h \in \calH} {L_{\distr,\context}(h)}$.
% Notice that the realizable setting is just a specific case when $L_{\distr,\context}( h^* )=0$.
% We assume that the hypothesis space $\calH$ is provided by the user, for example in case of MAX-SAT the user tells that the models can only be CNF.

The \emph{regret} refers to the cost incurred by using the solutions output by learned model $\program$ in place of those of the ground-truth $\program^*$~\cite{cesa2006prediction}.
%\samuel{where in this text? and what do you mean with this reference?}. \stefano{the whole book is about regret :-)}
Let
$
    \Optima(\context) \defeq \{ \vx \in \{0,1\}^n \,:\, h(\vx, \context) = 1 \}
$
be the set of context-specific optima of $h$ in context \context and $\Optima^*(\context)$ be the same for \add{ground-truth classifier} $h^*$.  Then, the context-specific and average regret are defined as follows:
\begin{definition}[Regret]
\label{def:regret}
Let $h^*$ be the true hypothesis and $\vx^* \in \Optima^*(\context)$.  The regret $\regret(\vx,\context)$ of using \add{assignment} $\vx$ in context \context is:
$$
    \regret(\vx,\context) \defeq
        \begin{cases}
            f_{\vw^*}(\vx^*) - f_{\vw^*}(\vx)
                & \text{if $\vx \models \context \land \phi(\vc)$}
            \\
            \wcr
                & \text{otherwise}
        \end{cases}
$$
\add{where $\wcr$ is the regret suffered by using an infeasible assignment.}
The average regret $\regret(h,\context)$ of a hypothesis $h$ in context \context is then the average regret of its optima, i.e.,
$$
    \regret(h,\context) \defeq \frac{1}{|\Optima(\context)|} \sum_{\vx \in \Optima(\context)} \regret(\vx, \context)
$$
\end{definition}

Hence, the average regret boils down to the average difference in quality between the \add{assignments} classified as positive by $h$ and by $h^*$. 

Notice that our definition of regret covers cases in which the learned model outputs infeasible \add{assignments}, \add{in which case the regret is $\wcr$.  The next theorem depends on the assumption that $\wcr$ is finite}, which is however reasonable in many applications.  For instance, if the learned model produces a schedule in which one nurse too many is asked to go to work, \add{the regret is roughly equivalent to the hourly pay of that nurse}.

As long as $\wcr < \infty$, the next result links the risk of a MAX-SAT classifier and the regret of the corresponding MAX-SAT model:

\begin{restatable}{theorem}{ThmRegretBound}
\label{thm:RegretBound}
Let $\minprob = \min_{\vx:D(\vx\mid\context) > 0} D(\vx\mid\context)$ and $\wcr$ be the regret associated to infeasible \add{assignments}. Then, for any $h \in \calH$ and $\context \in \Contexts^*$, it holds that:
$$
    \regret(h,\context) \le \frac{\|\vw^*\|_1 + \wcr}{\minprob |\Optima(\context)|} L_{\distr,\context}(h)
$$
This inequality holds for both realizable and agnostic settings\footnote{This bound improves on the one provided by Kumar et al.~\cite{kumar2019learning} by a constant factor.}.
\end{restatable}

As a consequence, minimizing the (empirical) risk of a MAX-SAT classifier entails minimizing the regret of the corresponding MAX-SAT model.  

Clearly, in applications in which infeasible \add{assignments} can violate societal or safety requirements, $\wcr$ may be infinite and the bound in Theorem~\ref{thm:RegretBound} becomes void.  This is expected, as in these conditions no worst-case guarantees can be given about the quality of the solutions output by \emph{any} learned model.  This essentially shows that, understandably, extra care must be taken when applying constraint learning to high-stakes domains.

\subsection{Generalizing across Contexts}
\label{sec:intercontext}

So far we have only considered learning and \add{solving} within a given context \context.  However, learned models are likely going to be used in previously unobserved contexts.  Next, we show that MAX-SAT classifiers learned from context-specific data do generalize, under mild assumptions, to any given target context $\context_\text{t}$ -- potentially distinct from the ones observed in the data set -- and therefore have low regret in $\context_\text{t}$ too.

Intuitively, in order for generalization to occur, the observed contexts have to be collectively \emph{representative} of the target one.
As an example, in nurse rostering, if all the example schedules were collected during a festive period in which most nurses are on holiday, it is unlikely that the learned model generalizes to non-festive periods in which most nurses are available.
More formally, we say that \add{a set of contexts} \Contexts is representative if whenever $h(\cdot,\context_\text{t})$ makes a mistake, there is at least a context in \Contexts that ``catches'' that mistake:\footnote{This definition generalizes the one in Kumar et al.~\cite{kumar2019learning} to arbitrary target contexts $\context_\text{t}$, and as a consequence Theorem~\ref{thm:lastone} is not limited to the top context $\context_\text{t} = \top$ anymore, \add{here $\top$ simply means global context, i.e., the absence of a context}}

%\stefano{TODO: $\errormatch$ is only violated by $\vx$'s that are optimal in $\context_1$ but are not optimal in $\context_2$ and apply geometric arguments.}

\begin{definition}[Representativeness]
\label{def:representative}
Fix a ground-truth classifier $h^* \in \calH$ and a hypothesis $h \in \calH$.  The formula $\errormatch(\context, \vx, \context')$ holds iff whenever $h$ misclassifies $\vx$ in context $\context$ it also misclassifies $\vx$ in context $\context'$, that is:
\begin{align*}
    \errormatch(\context, \vx, \context') =
        & \; (\vx \models \context \land \context')
    \\
        & \; \land (h(\vx, \context) \ne h^*(\vx, \context) \Rightarrow h(\vx, \context') \ne h^*(\vx, \context'))
\end{align*}
The number of contexts to which misclassification errors propagate to is:
$$
    \#(\context, \vx) = |\{ \context' \in \Contexts \,:\, \errormatch(\context, \vx, \context') \}|
$$
We say that a family of contexts \Contexts is representative for a target context $\context_\text{t}$ if and only if $\#(\context_\text{t}, \vx) > 0$ for all \add{assignments} $\vx$.
\end{definition}

\begin{figure}[tb]
    \centering

    \begin{tikzpicture}
        \begin{scope}[scale=1,shift={(1,0)}]

            % Header
            \draw (0.5, 2.85) node[above] {$X_1$};
            \draw (1.5, 2.85) node[above] {$X_2$};
            \draw (2.5, 2.85) node[above] {$X_3$};

            % Separator
            \draw[thick,black] (0.25, 2.9) -- (2.75, 2.9);

            % Rows
            \draw (0.5, 2.45) node[above] {$0$};
            \draw (1.5, 2.45) node[above] {$0$};
            \draw (2.5, 2.45) node[above] {$0$};

            \draw (0.5, 2.1) node[above] {$0$};
            \draw (1.5, 2.1) node[above] {$0$};
            \draw (2.5, 2.1) node[above] {$1$};

            \draw (0.5, 1.75) node[above] {$0$};
            \draw (1.5, 1.75) node[above] {$1$};
            \draw (2.5, 1.75) node[above] {$0$};

            \draw (0.5, 1.4) node[above] {$0$};
            \draw (1.5, 1.4) node[above] {$1$};
            \draw (2.5, 1.4) node[above] {$1$};

            \draw (0.5, 1.05) node[above] {$1$};
            \draw (1.5, 1.05) node[above] {$1$};
            \draw (2.5, 1.05) node[above] {$1$};

            \draw (0.5, 0.7) node[above] {$1$};
            \draw (1.5, 0.7) node[above] {$1$};
            \draw (2.5, 0.7) node[above] {$0$};

            \draw (0.5, 0.35) node[above] {$1$};
            \draw (1.5, 0.35) node[above] {$0$};
            \draw (2.5, 0.35) node[above] {$1$};

            \draw (0.5, 0) node[above] {$1$};
            \draw (1.5, 0) node[above] {$0$};
            \draw (2.5, 0) node[above] {$0$};

            % Separator
            \draw[thick,black] (0.25, 0) -- (2.75, 0);

            % psi_1
            \draw [decorate, decoration={brace,amplitude=10pt}, color=blue]
                (0.2, 0.1) -- (0.2, 1.45)
                node [black, midway, xshift=-0.6cm, color=blue] {$\context_1$};

            % psi_2
            \draw [decorate, decoration={brace,amplitude=10pt}, color=violet]
                (0.2, 1.5) -- (0.2, 2.8)
                node [black, midway, xshift=-0.6cm, color=violet] {$\context_2$};

            % psi_3
            \draw [decorate, decoration={brace,amplitude=10pt,mirror}, color=red]
                (2.8, 1.15) -- (2.8, 1.8)
                node [black, midway, xshift=+0.6cm, color=red] {$\context_3$};

        \end{scope}

    \end{tikzpicture}

    \caption{\label{fig:rep_context} Example of representative contexts: $\context_1 = X_1$ (blue), $\context_2 = \neg X_1$ (violet), $\context_3 = X_2 \land X_3$ (red) are representative for $\Phi=\{X_1,X_2,X_3,\neg X_1,\neg X_2,\neg X_3\}$ and $\vw^*=\{1,1,1,0,0,0\}$; details in the text.  (Best viewed in color.)}

\end{figure}

To see why representativeness is necessary, consider a ground truth model $\vc^* = (0, 0, 0, 0, 0, 0)$ and $\vw^* = (1, 1, 1, 0, 0, 0)$ with:
$$
    \Phi = \{X_1, X_2, X_3, \neg X_1, \neg X_2, \neg X_3\}
$$
$$
    \context_1 = X_1, \quad \context_2 = \neg X_1, \quad \context_3 = X_2 \land X_3, \quad \context_t = \top
$$
See Figure~\ref{fig:rep_context} for an illustration. Notice that under $\context_1$ the optimum is $(1, 1, 1)$ with value $3$, while under $\context_2$ the optimum is $(0, 1, 1)$ with value $2$.  Now, take a hypothesis~$h$ that misclassifies $\vx = (0, 1, 1)$ as a global optimum, \add{i.e., optimum in the target context $\context_t$}.  The error does neither show up in $\context_1$ (because $\vx$ does not lie in it) nor in $\context_2$ (because $\vx$ is indeed optimal in it).  In other words, $\errormatch(\context_\text{t}, \vx, \context_1)$ and $\errormatch(\context_\text{t}, \vx, \context_2)$ do not hold and $\#(\context_\text{t}, \vx) = 0$.  Therefore, a classifier with arbitrarily low risk on both $\context_1$ and $\context_2$ may still misclassify $\vx$.  Requiring that $h$ performs well also on $\context_3$ fixes this issue, because $\vx$ does lie in $\context_3$ but is not optimal in it -- and $\errormatch(\context_\text{t}, \vx, \context_3)$ holds.

Crucially, if \Contexts is representative and $h$ has low risk on all observed contexts in \Contexts, then it performs well in the target context too:

\begin{restatable}{lemma}{mixtotop}
\label{thm:mixtotop}
Let $\Contexts$ be representative for $\context_\text{t}$, and $\distr(\vx\mid\context) > 0$ for every $\context \in \Contexts$ and $\vx \models \context$.  Then there exist finite constants $\beta_\context \ge 0$ (that only depend on \Contexts and $D$) such that for every $h \in \calH$:
$$
    \textstyle
    L_{D,\context_\text{t}}(h) \le \sum_{\context \in \Contexts} \beta_\context L_{D,\context}(h)
$$
\end{restatable}

We are now ready to state our main result:
\begin{restatable}{theorem}{lastone}
\label{thm:lastone}
For any \Contexts and \distr that satisfy the conditions of Lemma~\ref{thm:mixtotop} and for every $\epsilon$, $\delta \in (0, 1)$, there exist integers $t_\context(\epsilon, \delta)$ for $\context \in \Contexts$ such that, if \dataset contains at least $t_\context(\epsilon,\delta)$ context-specific examples for every $\context \in \Contexts$, then any hypothesis $h$ with minimal empirical risk on \dataset satisfies:
$$
    \prob(L_{\distr,\context_\text{t}}(h) > \min_{h' \in \calH}\sum_{\context \in \Contexts}L_{\dataset,\context}(h') + \epsilon) < \delta
$$
Notice that, in the realizable setting this reduces to:
$$
    \prob(L_{\distr,\context_\text{t}}(h) > \epsilon) < \delta
$$
\end{restatable}

Summarizing, as long as the observed contexts are representative and there are enough examples, ERM learns a low-risk MAX-SAT model that has low regret in both the observed and in the target context with high probability.

\section{Two Implementations}
\label{sec:hassle}

We are finally ready to present our implementations of MAX-SAT learning.  Both are based on Empirical Risk Minimization, which, given $\vX$, $\Phi$, and a context-specific data set \dataset encompassing contexts \Contexts, amounts to searching for a MAX-SAT model \program that minimizes $\sum_{\context \in \Contexts}L_{\dataset,\context}(h)$.  This equates to solving the following optimization problem:
\begin{align}
    \mathrm{find}
        & \quad \vc \in \{0, 1\}^m, \vw \in [-1, 1]^m \label{eq:learning-objective}
    \\
    \text{s.t.}
        & \quad y_k \Leftrightarrow \left( \vx_k \in \argmax_{\vx' \,\models\, \phi(\vc) \,\land\, \context_k} f_{\vw}(\vx') \right)
        & k = 1, \ldots, s \label{eq:learning-constraint}
\end{align}
\add{Intuitively, Eq.~\ref{eq:learning-objective} searches for two vectors $\vc$ and $\vw$ that encode a MAX-SAT model, which is constrained by Eq.~\ref{eq:learning-constraint} to classify all examples correctly.  More in detail, said model should predict all positive training examples to be both optimal (w.r.t. the learned objective $f_{\vw}$) and feasible (w.r.t. the learned hard constraints and the context $\phi(\vc) \land \context_k$) and all the negatives to be either sub-optimal or infeasible.}

Different solution strategies can be applied.  In the following, we introduce two different implementations, \methodmilp and \methodsls, based respectively on syntax-guided synthesis and on stochastic local search. For both the methods, we also assume the user to provide an upper bound on the number of constraints to be learnt. Some strategies can be used to avoid this, for example, start by learning a few constraints and then increase the number if the learnt model has a low accuracy, we can also learn different number of constraints in parallel and then pick one with the best accuracy. However we leave these extensions for the future work. Now we will discuss both \methodmilp and \methodsls in turn.

\subsection{MAX-SAT Learning with MILP}

Following the literature on syntax-guided synthesis~\cite{alur2018search}, we solve the ERM problem directly by encoding it in a suitable optimization formalism and then applying an appropriate solver.  This solution offers three main advantages.  The main one is that the hard problem of finding a suitable candidate is offloaded to the solver, in a declarative fashion.  In addition, using an exact solver guarantees that the empirical error of the learnt model is exactly zero, and will trigger a warning in cases where this is impossible. A third advantage is that, if the solver supports anytime execution, then a sub-optimal solution can be obtained by specifying a time budget.\footnote{Some supposedly anytime solvers are not anytime in practice, as they have a bootstrap stage in which they look for an initial solution and only afterwards attempt to improve upon it.  As a consequence, if the cutoff is shorter than the time taken to find this initial solution, then no candidate can be returned by the solver.  This is what happens in our experiments.}

However, a na\"ive encoding of Eq.~\ref{eq:learning-constraint} is non-trivial, because checking whether an example is positive (i.e., feasible and optimal) requires to solve a nested partial MAX-SAT \add{model}.  Furthermore, this has to be done $s$ times, once for each training example.  Instead of dealing with nesting, we find it more convenient to encode the problem into a mixed integer linear programming (MILP) problem, as shown in Figure~\ref{fig:milp_encoding}.  The encoding can be split into two parts.  First, the objective function (Eq~\ref{eq:obj}) maximizes a set of fresh per-context variables $\gamma_\context \in \bbR$ and together with Eq.~\ref{eq:xprime} computes a feasible optimum $\vx'_{\context_\ell}$ for every context $\context_\ell \in \Contexts$ appearing in the data.  Second, Eq.~\ref{eq:main} ensures that $\vc$ and $\vw$ are chosen so that every positive example $\vx_k$ is feasible and optimal in its own context $\context_k$ (using the context-specific optimum $\vx'_{\context_k}$ as a reference) and that no negative \add{assignment} is.  Once solved, the learned MAX-SAT model can be read off of $\vc$ and $\vw$.

\begin{figure*}[tb]
    \begin{align}
        \add{\argmax}_{\vc, \vw, \vgamma}
            & \; \textstyle \sum_{\context_\ell \in \Contexts} \gamma_{\context_\ell} 
            \label{eq:obj}
        \\
        \text{s.t.}
            & \; (\vx'_{\context_\ell} \models \phi(\vc) \land \context_\ell) \land (\gamma_{\context_\ell} \le f_{\vw}(\vx'_{\context_\ell}))
            & \; \context_\ell \in \Contexts \label{eq:xprime}
        \\
            & \; y_k \Leftrightarrow (\vx_k \models \phi(\vc) \land \context_k) \land (\gamma_{\context_k} \le f_{\vw}(\vx_k))
            & k = 1, \ldots, s \label{eq:main}
        \\
            & \vgamma \in \bbR^{|\Contexts|}, \; \vc \in \{0, 1\}^m, \; \vw \in [0, 1]^m \nonumber
    \end{align}
    \caption{\label{fig:milp_encoding} \methodmilp encodes the learning constraint (Eq.~\ref{eq:learning-constraint}) as a MILP problem.  A simplified encoding is shown here (the full encoding is given in~\ref{appendix:milp}). Eq.~\ref{eq:main} encodes the main constraint, i.e., an example is positive if-and-only-if it is both feasible and optimal.  Equations~\ref{eq:obj} and~\ref{eq:xprime} solve the inner MAX-SAT optimization problem ($\gamma_{\context_\ell}$ encodes the value of the optimal solution for context~$\context_\ell$).
    }
\end{figure*}

A couple of important remarks are in order.
The first one is that $\gamma_\context$ is always bounded thanks to Eq.~\ref{eq:xprime}.  In addition, for contexts that have at least one positive example, Eq.~\ref{eq:xprime} is superfluous and can be omitted for efficiency.

The second one is that the MILP encoding does not exactly correspond to the encoding in Eq.~\ref{eq:learning-constraint} -- it is, instead, a tight approximation.  The intuition is as follows.  To solve the nested MAX-SAT \add{model}, the MILP encoding uses variables~$\gamma_{\context_\ell}$ to represent the value of an optimal solution in a context~$\context_\ell$ according to the MAX-SAT \add{model} encoded by~$\vc$ and~$\vw$.  However, instead of individually maximizing the~$\gamma_{\context_\ell}$, it has to resort to maximizing the sum~$\sum_{\context_\ell} \gamma_{\context_\ell}$.  This summation introduces a dependency that allows the encoding to choose~$\gamma$ sub-optimally for some contexts -- allowing it to falsely label examples as positive that are not positive w.r.t.~$\vc$ and~$\vw$ -- in order to increase the~$\gamma$'s for different contexts.  A detailed example is given in~\ref{appendix:milp}.  A simple post-processing step can detect such sub-optimal~$\gamma$'s and an iterative procedure\footnote{In cases where falsely labeled positives are detected, an iterative procedure could be used to solve the problem to optimality:  For every context~$\context_\ell$ for which there exists an \add{assignment}~$\vx'$ with a value larger than a positive \add{assignment}~$\vx^+ \models \context_\ell$: $f_{\vw}(\vx') > f_{\vw}(\vx^+)$, a constraint: $\vx' \not \models \phi(\vc) \lor f_{\vw}(\vx') \leq f_{\vw}(\vx^+)$ is added to the encoding.  The extended encoding is solved and this procedure is repeated until all \add{assignments} are labeled correctly.} could be used to solve the problem, but our experimental results show that this mismatch occurs rarely in practice.  Indeed, in our experiments, the MILP encoding makes a mistake on the training set less than $1\%$ of the time.

\subsection{MAX-SAT Learning with Stochastic Local Search}

\methodmilp presented in the previous section is meant as a near-exact proof of concept implementation, and while it provides guarantees on the empirical risk of the learned model, it is not optimized for efficiency. Using an exact solver in \methodmilp also has a limitation. In an agnostic setting it will be unable to learn a model as the empirical error can not be reduced to zero. This also means that if the training data is noisy it might not be able to find a model. To overcome these limitations, we introduce an alternative implementation, \methodsls, that makes use of stochastic local search (SLS)~\cite{hoos2004stochastic} techniques.

Given a data set $\dataset$ and a set of candidate constraints $\Phi$, the aim of \methodsls is to find $\vc \in \{0, 1\}^m$ and $\vw \in [-1, 1]^m$, such that the corresponding classifier $h_{\vc,\vw}$ correctly classifies as many examples in $\dataset$ as possible, and it does so using SLS. The backbone of all SLS strategies is a heuristic search procedure that iteratively picks a promising candidate in the neighborhood of the current candidate, while injecting randomness in the search to escape local optima. Many SLS algorithms can be designed by changing the definition of the neighbourhood and the way in which a promising candidate is chosen. For \method-SLS we designed these choices based on ideas borrowed from WalkSAT algorithm and its variants. This choice is based on the simplicity as well as a performance comparison made with other alternatives by Hoos and Tsang~\cite{Hoos06localsearch}. We will discuss the variants of WalkSAT later in the section.

The pseudo-code for the WalkSAT version of \methodsls is shown in Algorithm~\ref{alg:hassleSLS}. It starts with a hypothesis $h_{\vc,\vw}$ sampled at random from $\calH$ (line~\ref{eq:slsinit}), and then iterates through three steps in a loop.  In the first step, a biased coin is tossed and with probability $p$ a random hypothesis is chosen as the next candidate (line~\ref{eq:slsrandom}), which is equivalent to a random restart.  If the check fails, a \emph{neighbourhood function} $\neighbour(h_{\vc,\vw}): \calH \rightarrow 2^{\calH}$, where $2^\calH$ is the power-set of $\calH$, is used to generate neighbours of the current candidate hypothesis, a \emph{scoring function} $\score(h_{\vc,\vw}): \calH \rightarrow \mathbb{R}$ assigns a score to each hypothesis in the neighborhood, and finally \add{a neighbour} with the highest score is picked as the next candidate, while keeping track of the best candidate model found so far.  The loop repeats until either a good enough hypothesis is found or an iteration or time budget is exhausted.
The score of a classifier $h_{\vc,\vw}$ is defined to be the accuracy on the training set, i.e., the number of correctly classified examples in the training set $\dataset$:
$$
    \score(h_{\vc,\vw})=\sum_{k=1}^s \Ind{h_{\vc,\vw}(x_{k},\context_{k}) = y_{k}}
$$
The choice of neighbourhood function is critical to the success of the learning procedure.  In the following, we discuss the choice made for \methodsls.

\begin{algorithm}[tb]
    \begin{algorithmic}[1]
        \Procedure{\methodsls}{$\calH$: candidates, $\dataset$: dataset, $p$: restart probability, $s$: cutoff score, $t$: cutoff time}
            \State $h \gets$ random candidate from $\calH$ \label{eq:slsinit}
            \State $h' \gets h$
            \While {$\score(h) < s \,\land\, \mathrm{runtime} < t$}
                \State \textbf{with probability $p$ do}
                    \State \qquad $h \gets$ random candidate from $\calH$ \Comment{random restart} \label{eq:slsrandom}
                \State \textbf{otherwise}
                    \State \qquad $h \gets \argmax_{h \in \neighbour(h')}\score(h)$ \Comment{depends on SLS strategy} \label{eq:slsargmax}
                \If{$\score(h) > \score(h')$}
                    \State $h' \gets h$ \Comment{track best-so-far}
                \EndIf
            \EndWhile
            \State \Return $h'$
        \EndProcedure
    \end{algorithmic}
    \caption{\label{alg:hassleSLS} The \methodsls WalkSAT algorithm.}
\end{algorithm}

%With this representation, the search space boils down to $\calH$ (see Definition~\ref{def:model}).  \stefano{@Mohit: $\gets$ I am not sure what this adds, sorry.}

\paragraph{Neighbourhood Function} Given a candidate hypothesis $h_{\vc,\vw}$ as input, the neighbourhood function $\neighbour(h_{\vc,\vw})$ determines the candidate models that can be reached in one search step. Notice that $\neighbour(h_{\vc,\vw})$ takes a hypothesis as input and gives a set of hypotheses as output.  The hypotheses $h_{\vc',\vw'} \in \neighbour(h_{\vc,\vw})$ can be obtained by either:
%As each hypothesis $h_{\vc,\vw}$ is uniquely represented by $(\vc,\vw)$, we will use $(\vc,\vw)$ and $h_{\vc,\vw}$ interchangeably below:

%   \stefano{from what I remember, an ideal neighborhood function should enable an SLS algorithm to quickly explore the space of candidates, i.e., to move quickly from one end to the other of the space, while producing small neighborhoods so that scoring them doesn't take ages.}
% To keep things simple and efficient, in \methodsls neighbors are generated by altering either $\vc$ or $\vw$ in a simple manner. As each hypothesis $h_{\vc,\vw}$ is uniquely represented by $(\vc,\vw)$, we will use $(c_1, \ldots, c_n, w_1, \ldots, w_n)$ to represent $h_{\vc,\vw}$. Using this notation we define neighbourhood function as follows:
% $(c'_1, \ldots, c'_n, w'_1, \ldots, w'_n) \in \neighbour(c_1, \ldots, c_n, w_1, \ldots, w_n)$
% In particular, a hypothesis $h_{\vc^{'},\vw^{'}}$ is a neighbor of the candidate $h_{\vc,\vw}$, written $h_{\vc',\vw'} \in \neighbour(h_{\vc,\vw})$, if exactly one of the following is true:
% Notice that the neighbourhood function can also be written as $\neighbour(c_1, \ldots, c_n, w_1, \ldots, w_n)$
%
\begin{itemize}

    \item Turning a soft constraint in $h_{\vc,\vw}$ into a hard constraint, that is, for some $j$ if $c_j = 0$ and $w_j \ne 0$ then $c'_j$ is set to $1$.  The corresponding weight $w'_j$ is set to $0$, for simplicity.
    
    \item Turning a hard constraint in $h_{\vc,\vw}$ into a soft constraint, that is, for some $j$ if $c_j = 1$ then $c'_j$ is set to $0$.  The corresponding weight $w'_j$ is set to $1$, for simplicity.
    
    \item Decreasing the weight of a soft constraint, that is, for some $j$ if $c_j = 0$ and $w_j \ne 0$ then $w'_j<w_j$, for simplicity we set $w'_j=w_j/2$.
    
    \item Increasing the weight of a soft constraint, that is, for some $j$ if $c_j = 0$ and $w_j \ne 0$ then $w'_j>w_j$, for simplicity we set $w'_j=(1+w_j)/2$.
    
    \item Making a minimal change in an existing constraint. As each constraint is a clause in our case, this means either 1) adding a literal to a clause, 2) removing a literal from a clause or 3) changing the sign of a literal in a clause. 
    % \begin{align*}(c_1,..,c'_i,..,c'_j,.., c_n, w_1,.., w'_i,..,w'_j,.., w_n) \in \neighbour(\vc,\vw)\; s.t. \\  \; c'_i = c_j \land w'_i = w_j \land c'_j = 0 \land w'_j = 0 \land |\phi_{i} - \phi_{j}|=1\end{align*}
    % $$
    %     \exists \; i \,:\, c'_i = 1 - c_i \,\land\, \forall j \ne i \,:\, c'_j = c_j
    % $$

    % \item The weight of a single soft constraint is either increased or decreased. To increase the weight we take the midpoint between the current weight and the maximum possible weight, which is $1$. Similarly, to decrease the weight we take the midpoint between the current weight and the minimum possible weight, which is $0$.
    % %
    % \begin{align*}
    %     \exists \; i \,:\,
    %         & \textstyle (c_i = 0 \; \land \; (w'_i = \frac{1}{2}(1 + w_i) \,\lor\, w'_i = \frac{1}{2}w_i))
    %     \\
    %         & \textstyle \land\, \forall \, j \ne i \,:\, w'_j = w_j
    %     % \\
    %     %     & \textstyle \land\, \vc' = \vc
    % \end{align*}

    % \item A single clause changes due to the addition, removal or sign switch of a literal:
    % %
    % \begin{align*}
    %     \exists \; i,j \,:\, 
    %     & \textstyle (c_{i},w_{i}) \ne (0, 0) \; \land \;  (c_{j},w_{j}) = (0, 0) \; \land \; (c'_{j},w'_{j}) = (c_{i},w_{i}) 
    % \\ 
    %     & \textstyle \land \; (c'_{i},w'_{i}) = (0, 0) \; \land \; |\phi_{i} - \phi_{j}|=1 
    % \\
    %     & \textstyle \land \, \forall \, k \ne i, j \,:\, c'_{k},w'_{k}=c_{k},w_{k} 
    % \end{align*}

\end{itemize}

Evaluating each hypothesis in $\neighbour(h_{\vc,\vw})$ requires solving an optimization problem, so a large neighbourhood can significantly impact the efficiency of the algorithm.
Moreover, notice that the size of the neighbourhood increases exponentially with the number of variables. Therefore, our algorithm won't scale if we evaluate each neighbour to select the next candidate.
Hence we designed a greedy approach to prune the neighbourhood, and evaluate only those models which show some promise of being better than the current candidate. To do so we take the following steps:
1) Randomly select one of the incorrectly classified examples by the current candidate,  
2) Evaluate only \add{those hypotheses} in $\neighbour(h_{\vc,\vw})$ which can make the selected example consistent.
Doing the first step involves finding the optimal value for the current candidate, i.e., solving just one optimization problem.
However, doing the second step is not so trivial, one obvious way would be to evaluate each candidate in $\neighbour(h_{\vc,\vw})$ on the selected example, but it defeats the purpose of not evaluating each candidate. To avoid this, we prune the neighbourhood based on the true and predicted label of the incorrectly classified example selected in step 1. For instance, if the true label is positive while the learnt model predicts it to be infeasible, we evaluate only those neighbours which make changes in the hard constraints which are not satisfied by the example. This makes sense, because only making such change can make the example feasible. See~\ref{appendix:neighbours} for the complete detail of how we prune the neighbourhood based on the true and predicted label.

\paragraph{WalkSAT variants}

In the WalkSAT implementation we used random restarts, which allows the search to get out of a local optimum, however doing a random restart does not utilise the progress made so far. This shortcoming is overcome by using \emph{tabu search}. It uses memory to prevent the search process from stagnating at a local optimum. The idea is to keep a short-term memory of most recently visited candidates and avoid revisiting them. 
WalkSAT has three extensions based on this idea: Novelty, Novelty$^+$ and Adaptive Novelty$^+$. Novelty keeps track of previous candidates and gives a preference to the novel ones. Novelty$^+$ also does the same but with a probability chooses a random neighbour.  Adaptive Novelty$^+$ takes it even further by changing the probability to choose a random neighbour:  if the improvement achieved in an iteration is high, it keeps the probability low and as the improvement decreases, it increases the probability. In application none of these is a clear winner, their relative performance varies with the problem. Hence we implemented each of the three variants along with the basic version given in Algorithm~\ref{alg:hassleSLS}. 
% Note that maximizing score in this case is equivalent to minimizing empirical risk, which means our learning algorithm is based on ERM. In the previous section we already proved the learnability results when using ERM as a learning algorithm, hence all the theoretical guarantees remain valid.
Through the experiments in the next section we analyze the accuracy of each method and answer various research questions.

\section{Experiments}
\label{sec:exp}

In this section, we empirically answer the following research questions:
\begin{description}

    \item[Q1] Does ERM succeed in acquiring good quality MAX-SAT models from contextual examples?

    \item[Q2] Among the four implementations of \methodsls, which SLS strategy performs the best and how does it compare to \methodmilp?

    \item[Q3] How does \methodsls scale as the complexity of the ground-truth model increases?

    \item[Q4] How good is our strategy of neighbourhood pruning compared to a naive implementation where each neighbour is evaluated? 

    \item[Q5] Is having both infeasible and sub-optimal negative examples essential to learn a good model? 

    \item[Q6] Can \methodsls handle noisy data?

    \item[Q7] \add{Is having contextual information essential to learn a good model?}

\end{description}
To this end, we implemented both \methodmilp and four different versions of \methodsls based on different SLS strategies, namely WalkSAT, Novelty, Novelty$^+$ and Adaptive Novelty$^+$.  Then we used these five implementations to recover both synthetic and benchmark ground truth MAX-SAT models of increasing complexity from contextual examples.  
\add{In the theory, to represent a MAX-SAT model, we use a vector of all constraints and a weight vector over all possible constraints, however, this is purely a theoretical concept. For the implementations
we restrict ourselves to a fixed number of constraints which are then implicitly (or lazily) generated
during the search (finding literals using MILP or manipulating constraints using SLS).}
The complete experimental setup can be found at:  \url{https://github.com/mohitKULeuven/HassleWithLocalSearch}

\subsection{Datasets}

The ground-truth \emph{synthetic models} were generated by enumerating all possible disjunctions of up to \add{$n/2$} literals of $n$ variables, and then sampling $\vc^*$ and $\vw^*$ at random so that they have exactly $m_\text{hard}$ hard and $m_\text{soft}$ soft constraints.  The weights of the soft constraints were sampled uniformly from $(0, 1]$.  To answer $\textbf{Q3}$, synthetic models were generated by varying the number of variables: $n = 8, 10, 12, 15$, hard constraints $m_\text{hard} = 2, 5, 10, 15$, and soft constraints $m_\text{soft} = 2, 5, 10, 15$.  Five random models were obtained for each \add{assignment} by changing the seed for randomization.

% Answering $\textbf{Q7}$ turned out to be more difficult than anticipated. Generating a uniformly random MAX-SAT model with large number of variables is itself a challenging task as it leads to exponentially large number of possibilities for constraints. So 
To validate our implementations on more realistic combinatorial \add{models}, we also chose five \emph{benchmark models} from a repository of phase-transition SAT instances~\cite{Gent1994TheSP, HoosStu}.\footnote{Taken from: \url{www.cs.ubc.ca/~hoos/SATLIB/benchm.html}}  Since benchmark models are meant to be used by solvers to test their performance, they are typically huge, with thousands of variables and constraints. We restricted the analysis to models with a reasonable number of variables, namely 20.  In addition, the set of constraints given in the benchmark models were reduced from 91 to 20 through random selection. Note that these benchmarks are SAT instances, hence we created MAX-SAT instances by converting half of the randomly selected constraints into soft constraints by adding random weights.

For each ground-truth model $\program^*$, a dataset \dataset was collected by first sampling $|\Contexts| = 25, 50, 100$ random contexts and then taking 2 positive and 2 negative examples from each context.  Contexts were chosen to be a \add{conjunction} of $n/2$ literals, for simplicity. \add{We also make sure that each context impacts the optimal solution of $\program^*$}.
%Ideally, contextual examples should be sampled randomly from each context.
\add{By negative example in a context $\context$, we refer to an example $\vx$ that satisfies the context $\context$, but is infeasible or sub-optimal with respect to the model $\program^*$.} Negative examples are generated through rejection sampling, making sure that half of them are sub-optimal while half are infeasible. Generating truly random positive examples, however, is not trivial. Hence, we first enumerated 10 times the required number of positive (optimal) examples by solving $\program^*$ under the given context using a MAX-SAT solver\footnote{Using the PySAT solver interface~\cite{ignatiev2018pysat}.} and then resorted to reservoir sampling~\cite{tille2011sampling} to obtain the required number of samples. For each model $\program^*$, we generated 5 different set of data by changing the seed of randomization.

\subsection{Performance Measures}

% Recall that, in order for ERM to work, it is necessary that the contexts for which data is available are representative of the target context, see Definition~\ref{def:representative}.  \add{Indeed, if the training contexts are not representative of the target context, then any learned MAX-SAT model can perform arbitrarily poorly. \stefano{not good enough!}}  To ensure \add{this does not occur} in our experiments, the performances of the learned models were measured in a target context
% \noindent The performance of the learned models was measured in the target context. To ensure that the representativeness property in Definition~\ref{def:representative}, target context was chosen to be the union (disjunction) of the contexts used for sampling training data. 
The performance of the learned models were measured in terms of \emph{score} (that is, training set accuracy) as well as \emph{accuracy} in the \add{global} context, \emph{regret}, and \emph{infeasibility}. The score is the optimization criterion used (either implicitly or explicitly) by \methodmilp and \methodsls, and it was used here to capture how well the various maximize their objective.  The accuracy measures the performance of the acquired MAX-SAT model $\program$ (when viewed as a classifier) in the \add{global} context, while regret measures the quality of its optima in the \add{global} context.  Finally, infeasibility is defined as the percentage of optimal solutions in $\program$ which are infeasible in $\program^*$.

\subsubsection{Computing accuracy and infeasibility using model counting}

Calculating these performance measures is not trivial.
To exactly measure accuracy and infeasibility we rely on a technique called Model Counting (MC or \#SAT).  MC is a technique to count the feasible solutions of a propositional formula. If $\mc(\theta)$ denotes the model count of a formula~$\theta$ over Boolean variables~$\vX$, then we can compute the accuracy of a learned formula $\theta^L$ compared to an optimal formula~$\theta^*$ by computing:
$$
    \mathrm{acc}_{\theta^*}(\theta^L) =
    \frac{\mc(\theta^L \land \theta^*)}{2^{|\vX|}} + 
    \frac{\mc(\lnot \theta^L \land \lnot \theta^*)}{2^{|\vX|}}
$$
In our case, however, we learn MAX-SAT programs with both hard and soft constraints, and we are interested in both the accuracy of the optimal solutions and the infeasibility ratio.  To use model counting we need to compute a logic formula that describes the optimal solutions of a MAX-SAT program~$\program$. Let us denote the hard constraints of~$\program$ as $\program_H$, \add{the weights of~$\program$ as $\vw$}, an optimal solution as~$x^*$ and its value according to~$\program$ as \add{$v^* = f_{\vw}(x^*)$}.
To describe the optimal solutions using a formula $\theta_{\program}$, we find all subsets~$S_i$ of soft constraints of~$\program$ for which it holds that the sum of the weights of the soft constraints in~$S_i$ sums up to~$v^*$.  We then obtain~$\theta_\program = \program_H \land (\bigvee_{S_i}  \bigwedge_{\theta_s \in S_i} \theta_s)$, i.e., a solution to~$\theta_\program$ must satisfy the hard constraints and satisfy one of the sets of soft constraints that achieves the optimal score.  Using $\theta_{\program}$ and $\theta_{\program^*}$ we can now compute the accuracy as above and the infeasibility as: $$\frac{\mc(\theta_{\program} \land \lnot \program^*_H)}{\mc(\theta_{\program})}$$

\subsubsection{Computing regret through sampling}
Computing regret is even more tricky, notice that in Definition~\ref{def:regret} regret for infeasible examples is a constant. In our experiments, instead of fixing this value we compute regret only for feasible examples. This makes sense as we already compute infeasibility, which gives us the percentage of examples which are infeasible. Hence in all the experiments regret and infeasibility should be seen together to get the complete picture.
To calculate regret, we generated 1000 positive examples using $\program$ such that these examples are also feasible in $\program^*$ and then computed regret using Definition~\ref{def:regret}. To simplify the comparison, the regret was normalized to $[0, 1]$ by dividing it by $f_{\vw^*}(\vx^*)$, where $\vx^*$ is an optimum of $\program^*$.
%
% For the benchmark instances, which are larger, weighted model counting failed to calculate accuracy and infeasibility in reasonable time, so we computed these using a Monte Carlo approximation. 
% Namely, we sampled $1000$ examples from $\program$.  Accuracy was then computed as the percentage of the sampled examples which are correctly classified. 
We split the discussion of results in three parts, first we see the performance on synthetic data, then on noisy data and then finally on benchmark instances.

%We measure the performance of $\method$ to see if $\program$ generalises to the global context as well as to other contexts not seen during training. For the first part, we generate examples in the global context for $\program$ to calculate precision and regret, while for the second part we randomly choose some contexts not used in training and generate context-specific examples to measure performance.
\subsection{Results  on Synthetic Data}

For each experiment we randomly restart the local search if we get stuck at a candidate for more than a quarter of the cutoff time, which is the maximum time an algorithm is allowed to run before producing a learnt model. The probability of randomly selecting a neighbour for Novelty$^+$ and Adaptive Novelty$^+$ is $0.1$. 
\add{Note that here 0.1 is the probability of picking a random neighbour from the neighbourhood, while the probability p in Algorithm~\ref{alg:hassleSLS} is the probability of picking any random candidate, not necessarily in the neighbourhood.}
The other parameters for Adaptive Novelty$^+$ has been assigned the same values as in the experiments of the paper introducing this method~\cite{adaptiveNovelty}.

\subsubsection*{\textbf{Q1} Does ERM succeed in acquiring good quality MAX-SAT models from contextual examples?} 

\begin{figure*}[tb]
\vspace*{-1cm}
    \centering
        \includegraphics[width=\linewidth]{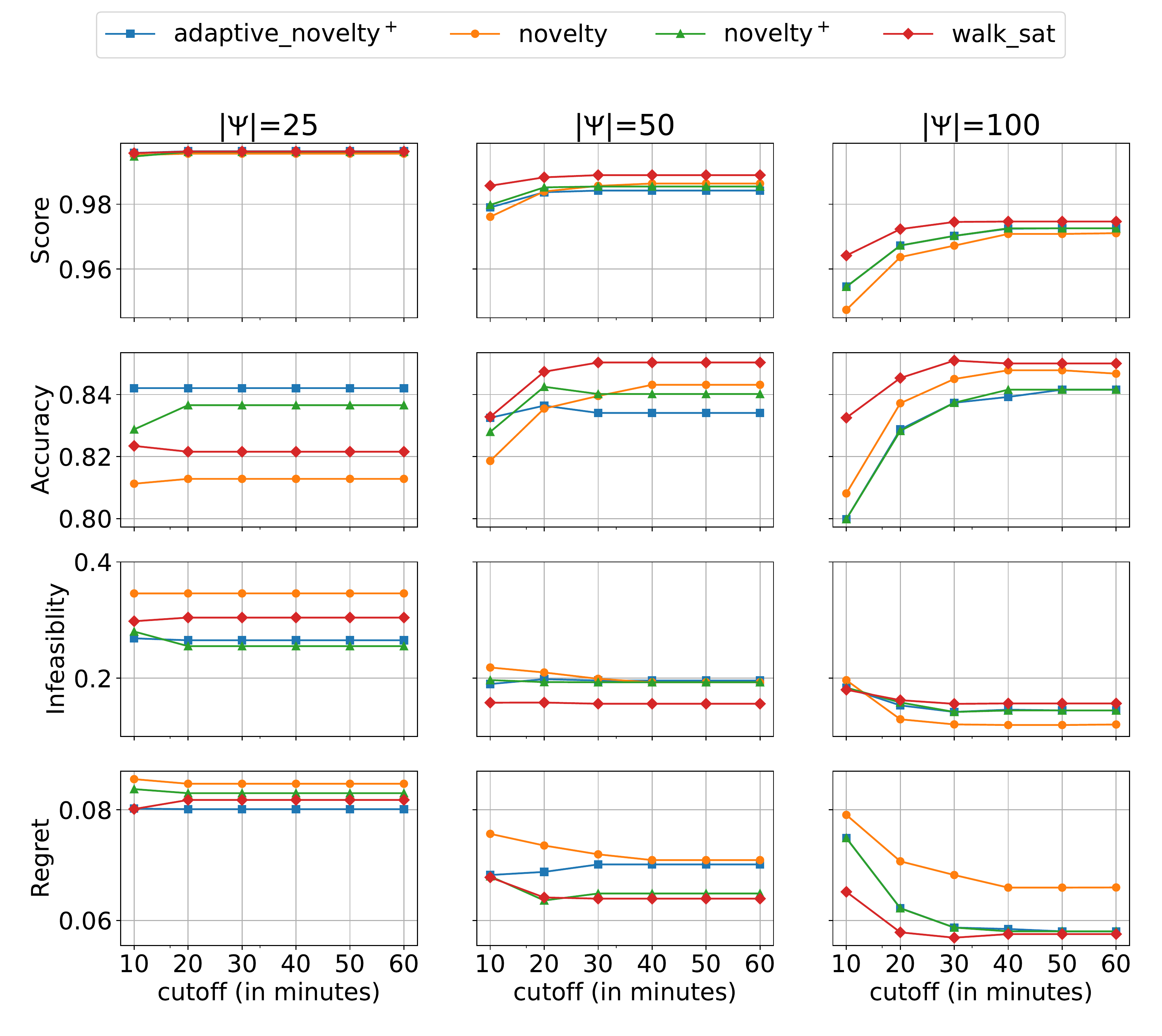}
        \caption{Performance evaluation on synthetic data. Performance of \methodsls improves as we increase the size of the training set by increasing $|\Contexts|$, with WalkSAT showing the best performance among all the implementations.
        }
    \label{fig:synthetic}
\end{figure*}

\begin{figure*}[tb]
\vspace*{-1cm}
\hspace{1.3 cm}
\begin{minipage}[b]{0.4\linewidth}
\includegraphics[width=\linewidth]{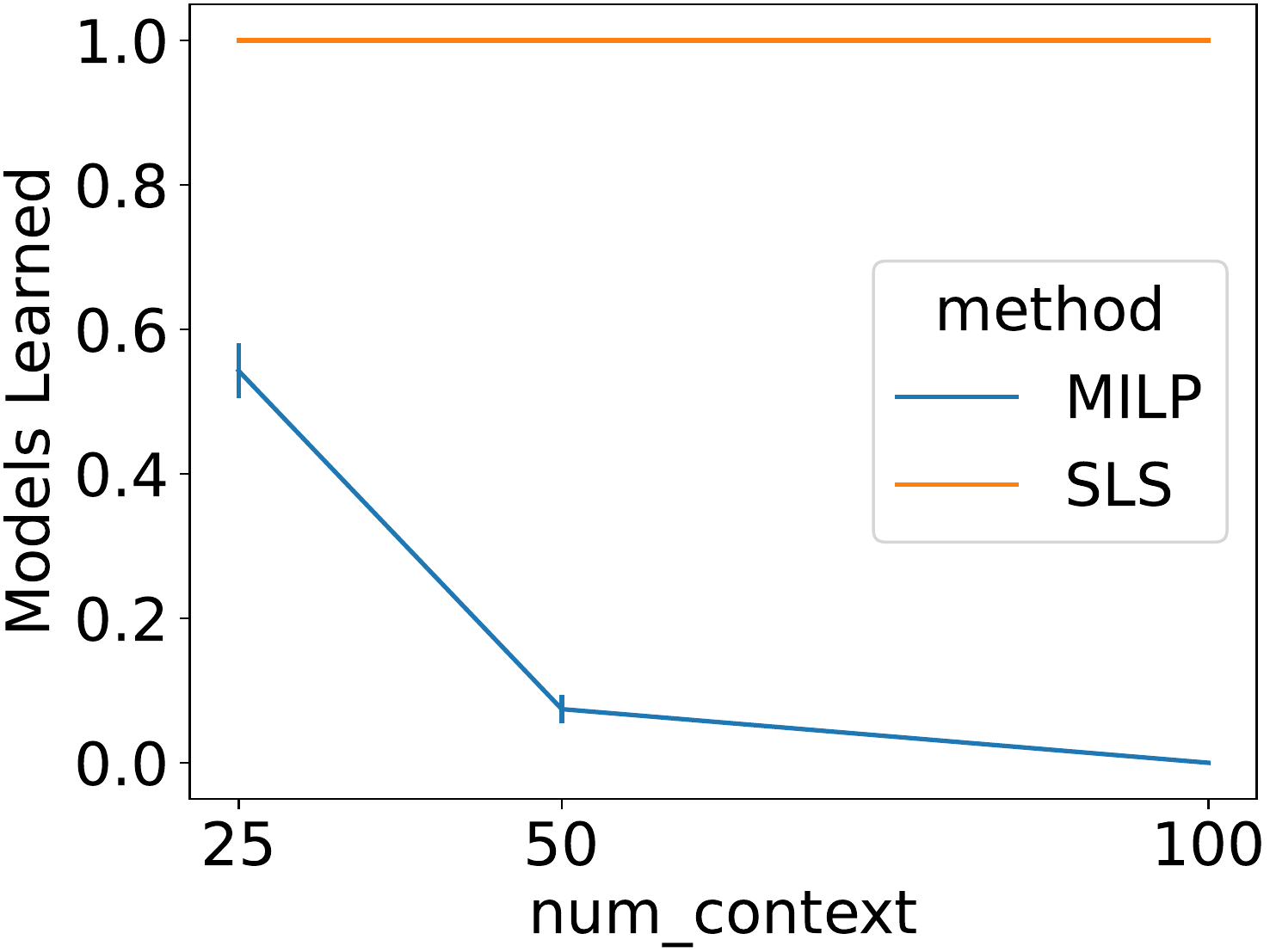}
\end{minipage}
\hspace{0.2 cm}
\begin{minipage}[b]{0.2\linewidth}
% \captionsetup{type=table} %% tell latex to change to table
% \begin{table*}[tb]
    \begin{footnotesize}
    \begin{tabular}[b]{ l| c| c}
        \toprule
            & \textbf{MILP}
            & \textbf{SLS}
        \\
        \midrule
        \textbf{Score} & $91.8$ & $98.7$ \\ 
        \textbf{Accuracy} & $74.7$ & $83.4$ \\ 
        \textbf{Infeasibility} & $50.1$ & $21.0$  \\ 
        \textbf{Regret} & $9.4$ & $7.3$  \\  
        \bottomrule
    \end{tabular}
    \end{footnotesize}
\end{minipage}
    \caption{\label{fig:milpvssls}  Left plot shows the percentage of experiments where the algorithm was able to learn a model within the specified cutoff time of 1 hour. As the size of training set increases, \methodmilp fails to learn any model, however \methodsls learns a model in each case. On the right, we compare the performance for the cases where \methodmilp learns a model, and even in those cases \methodsls clearly outperforms \methodmilp.}
% \end{table*}
\end{figure*}

We report the performance of all implementations on synthetic data in Figure~\ref{fig:synthetic}. Different columns in the figure corresponds to different number of contexts used to sample training data. From each context 2 positive and 2 negative examples were sampled, so increasing the number of contexts leads to larger training set.
On the $x$-axis of each graph we have cutoff time in minutes. Each row corresponds to different performance measure plotted on the $y$-axis. We can make the following observations from these plots: 
1) \add{Increasing the size of the training set leads to a better overall performance (3rd column) proportionally to the cutoff;}
2) When using 100 contexts with WalkSAT version of \methodsls, we achieve excellent performance: \add{$85\%$ accuracy,  $15\%$ infeasibility, and less than $6\%$ regret}. These numbers clearly show that ERM can learn high-quality MAX-SAT models using only contextual examples, thus answering $\textbf{Q1}$ in the affirmative.

% 1) Depending on the cutoff, \methodmilp either learns a model with $100\%$ score (1st row, 1st column) or nothing at all (line missing in the 2nd and 3rd column);

\subsubsection*{\textbf{Q2} Among the four implementations of \methodsls, which SLS strategy performs the best and how does it compare to \methodmilp?}

\add{
Looking at Figure~\ref{fig:synthetic}, it is clear that the overall performance of WalkSAT is better than the other implementations across all performance metrics.  For this reason, we focused on the WalkSAT version of \methodsls in the following experiments and will refer to it simply as \methodsls, unless stated otherwise.
Next, we compared \methodmilp and \methodsls to measure their scalability as well as quality. The results are shown in Figure~\ref{fig:milpvssls}.
In the left plot we show the percentage of cases across all experiments where \methodmilp and \methodsls were able to learn a model in the provided cutoff time of 1 hour.
As expected, \methodmilp fails to learn models due to the cutoff when we increase the size of the training set, while \methodsls learns a model in every case. Even in the cases where \methodmilp learns a model (Table in Figure~\ref{fig:milpvssls}), the performance of \methodsls is much better. Hence, \methodsls outperforms \methodmilp both in terms of quality and scalability.
}

\subsubsection*{\textbf{Q3} How does \methodsls scale as the complexity of the ground-truth model increases?}
\begin{figure*}[tb]
    \centering 
        \includegraphics[width=\linewidth]{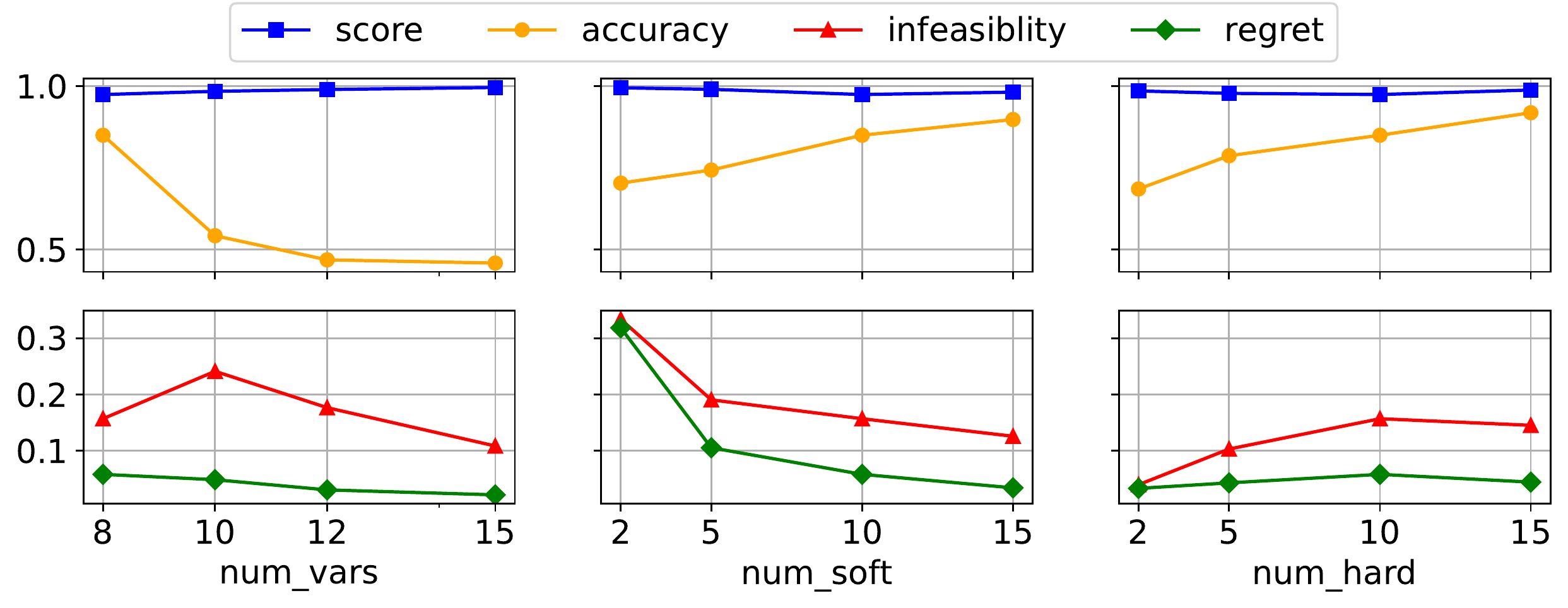}
    \caption{These results were generated by fixing $|\Contexts| = 100$ and with a cutoff of 60 minutes.
    }
    \label{fig:scalability}
\end{figure*}

To answer $\textbf{Q3}$, we see the change in performance by increasing the complexity of the target model, which is achieved by increasing the number of variables or the number of hard and soft constraints in $\program^*$.  We report the results for $100$ input contexts.  The results are shown in Figure~\ref{fig:scalability}. 
\add{When increasing the number of variables, we see a significant dip in the accuracy, this is due to the fact that the size of the feasible region increases exponentially, making it harder to learn it accurately with a small training set. However, the good thing is that the infeasibility and regret do not suffer, thus the learned model can still be used to generate good quality solutions.
Increasing the number of constraints, on the other hand, decreases the size of the feasible region, thus the accuracy improves when we increase either the number of soft constraints (2nd plot) or hard constraints (3rd plot). This also leads to better infeasibility and regret in the 2nd plot, however when increasing the number of hard constraints we observe slight increase in infeasibility, while regret does not show any significant change.} However, since increasing the number of contexts and the cutoff time improves the performance, as stressed by our theory and shown in Figure~\ref{fig:synthetic}, we expect that allowing more contexts, examples, and time to the learner would be enough to improve performance in these more complex models.

\subsubsection*{\textbf{Q4} How good is our strategy of neighbourhood pruning compared to a naive implementation where each neighbour is evaluated?}
\add{To answer $\textbf{Q4}$, we compared \methodsls with a naive version where we do not prune the neighbourhood. When the number of contexts is 50 and cutoff time is 1 hour, \methodsls explores $56\%$ less neighbours on average and achieves more than $9\%$ improvement in the score, from $89\%$ to $97\%$. This further translates to $73\%$ drop in infeasibility from $57\%$ to $15\%$ and $36\%$ drop in regret from $8.9\%$ to $5.7\%$. Most importantly \methodsls achieves these improvements in just $15$ minutes on average, which is much less compared to $35$ minutes taken by the naive version.  This allows us to confidently answer \textbf{Q4} in the affirmative.}

\subsubsection*{\textbf{Q5} Is having both infeasible and sub-optimal negative examples essential to learn a good model?}
To investigate the relative impact of sub-optimal and infeasible negative examples, we varied the percentage of the two in the data. We created three scenarios:  in the first one all negative examples are infeasible, in the second they are sub-optimal, and in the last one exactly half are sub-optimal and half are infeasible. The performance of \methodsls in these scenarios is shown in Table~\ref{tab:result_neg_type}. 
\add{The first row clearly shows that the absence of sub-optimal examples in the training set leads to a very inaccurate model, with the accuracy as low as $53\%$ and infeasibility as high as $49\%$. The second row shows that sub-optimal examples are essential to learn a good model, as the performance improves drastically compared to the first row. In the third row, the performance remains similar to the case when we use only sub-optimal examples, this suggests that infeasible negative examples are not that informative.}

\begin{table}[tb]
    \centering
    \ra{1}
    \begin{footnotesize}
    % \scriptsize
    \begin{tabular}{ m{2.0cm} m{1.0cm} m{2cm} m{2.5cm} m{2cm} }
        \toprule
        \textbf{Neg Data}
            & \textbf{Score}
            & \textbf{Accuracy}
            & \textbf{Infeasibility}
            & \textbf{Regret}
        \\
        \midrule 
        infeasible & $100$ & $53.0 \pm 2.0$ & $49.2 \pm 4.2$ & $13.6 \pm 0.8$ \\ 
        sub-optimal & $96.7$ & $84.8 \pm 1.9$ & $15.0 \pm 2.1$ & $6.4 \pm 0.9$ \\ 
        both & $97.5$ & $85.0 \pm 2.0$ & $15.6 \pm 1.6$ & $5.7 \pm 0.8$ \\
        % \midrule 
        % infeasible & $100$ & $10.6 \pm 2.4$ & $72.8 \pm 1.1$ & $14.3 \pm 0.6$ \\ 
        % sub-optimal & $93.7$ & $89.7 \pm 2.5$ & $21.1 \pm 4.8$ & $4.9 \pm 1.1$ \\ 
        % both & $95.5$ & $78.2 \pm 4.1$ & $30.0 \pm 6.6$ & $6.6 \pm 1.0$ \\
        \bottomrule
    \end{tabular}
    \end{footnotesize}
    \caption{Having sub-optimal examples is crucial to learn a good model. While, infeasible examples do not seem to be very informative, thus having only sub-optimal examples would also work to get a good model.}
    \label{tab:result_neg_type}
\end{table}

\subsection{Results on Noisy Data}

In this section, we examine the performance of \methodsls on noisy data.  We added noise in the synthetic data using the \emph{classification noise process}~\cite{angluin1988learning}, i.e., by randomly flipping the labels of each training example with some probability $p$.  One advantage of using \methodsls is that -- in contrast to \methodmilp -- it outputs a model even when the training set is inconsistent, i.e., no candidate hypothesis has empirical error zero.  Hence, it is suitable for noisy data too.

To generate the training data with noise, we picked the synthetic models generated for the earlier experiments with 8 variables, 10 hard and 10 soft constraints. We generated 100 random contexts to sample 2 positive and 2 negative examples from each. The label for each example was then randomly flipped with probability $p=0.05,0.1,0.2$.  

\subsubsection*{\textbf{Q6} Can \methodsls handle noisy data?}
The results for \methodsls are shown in Table~\ref{tab:result_noise}.  The first row corresponds to performance when there is no noise in the data.  In the subsequent rows, as $p$ increases, the performance of the learnt model decreases across all the performance metrics. \add{However, even with $p$ as high as $0.2$ which entails that around $20\%$ examples in the training set are mislabeled, we achieve $\sim 77\%$ accuracy, $\sim 32\%$ infeasibility and $\sim 9\%$ regret. Hence $\textbf{Q6}$ can be answered in the affirmative.}

\begin{table}[tb]
    \centering
    \ra{1}
    \begin{footnotesize}
    % \scriptsize
    \begin{tabular}{ m{1.5cm} | m{1.5cm}| m{2cm} | m{2.5cm} | m{2cm} }
        \toprule
        \textbf{Noise \%}
            & \textbf{Score}
            & \textbf{Accuracy}
            & \textbf{Infeasibility}
            & \textbf{Regret}
        \\
        \midrule
        $0$ & $97.5$ & $85.0 \pm 2.0$ & $15.7 \pm 1.6$ & $5.6 \pm 0.8$ \\ 
        $5$ & $91.5$ & $81.7 \pm 2.3$ & $18.9 \pm 2.5$ & $7.6 \pm 0.8$ \\ 
        $10$ & $87.1$ & $79.1 \pm 2.4$ & $28.1 \pm 3.1$ & $7.7 \pm 0.8$ \\ 
        $20$ & $78.6$ & $77.1 \pm 2.6$ & $31.9 \pm 3.3$ & $9.4 \pm 0.8$ \\  
        \bottomrule
    \end{tabular}
    \end{footnotesize}
    \caption{Performance of \methodsls on noisy data set. The first column specifies the probability of noise in the data, the other columns report performance metrics of the learned model.
    % As expected the performance decreases across all the metrics as the probability of noise in the data increases. However even with probability of noise as high as 0.2, we achieve $68\%$ accuracy with infeasibility $\sim 13\%$ and regret around $2\%$. 
    }
    \label{tab:result_noise}
\end{table}

\begin{figure*}[tb]
    \vspace*{-1cm}
    \centering
        \includegraphics[width=\linewidth]{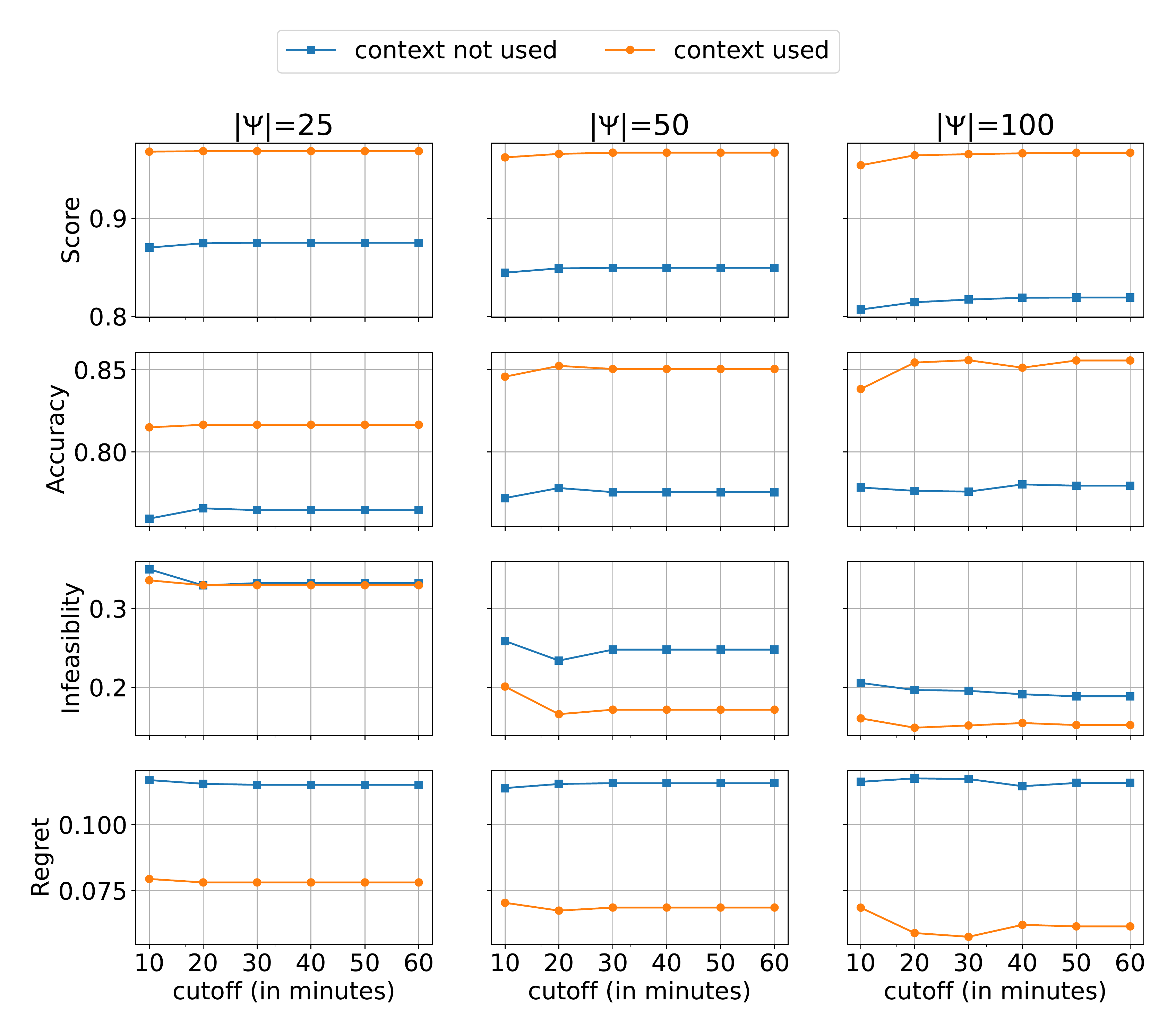}
        \caption{Performance evaluation on synthetic data. Performance across all metrics suffer if the contextual information is not used while learning.
        }
    \label{fig:synthetic_context}
\end{figure*}

\subsubsection*{\textbf{Q7} Is having contextual information essential to learn a good model?}

\add{To investigate the importance of contexts while learning a model, we designed experiments where we created a data set of contextual examples, but removed the information about the contexts. This replicates the learning setting for most of the constraint learning approaches, as they do not utilise the information about the context while learning. We learn model using this data and compare it with the case where we utilise the contextual information.
The results are shown in Figure~\ref{fig:synthetic_context}, as expected, using contextual information while learning improves the performance across all metrics. On an average, regret decreases by more than $\sim 50\%$ when using contextual information, while infeasibility decreases by $\sim 25\%$. These numbers clearly show that contextual information is quite essential to learn a good model.}

\subsection{Results on Benchmark Models}

\begin{table}[tb]
    \centering
    \ra{1}
    \begin{footnotesize}
    % \scriptsize
    \begin{tabular}{ m{1.5cm} | m{0.9cm}| m{1.5cm} | m{1.35cm} | m{1.35cm} | m{1.35cm}  | m{1.35cm} | m{0.8cm}}
         \toprule
         \textbf{Contexts}
            & \textbf{Score}
            & \textbf{Accuracy}
            & \multicolumn{2}{c|}{\textbf{Infeasibility}}
            & \multicolumn{2}{c|}{\textbf{Regret}}
            & \textbf{Time}
        \\
            & \multicolumn{1}{c|}{}
            & \multicolumn{1}{c|}{}
            & \textbf{Hassle}
            & \textbf{Random}
            & \textbf{Hassle}
            & \textbf{Random}
            & (hrs)
        \\
         \midrule
        %  $100$ & $99.9$ & $83 \pm 1.8$ & $54 \pm 3.6$ & $74 \pm 0.5$ & $8 \pm 1.2$ & $13 \pm 0.3$ & $1.7$ \\
         $250$ & $97.5$ & $72.6 \pm 2.6$ & $40.0 \pm 3.5$ & $74 \pm 0.5$ & $8.4 \pm 0.6$ & $12.8 \pm 0.4$ & $4.5$ \\
         $500$ & $96.2$ & $80.8 \pm 2.1$ & $26.8 \pm 3.2$ & $74 \pm 0.5$ & $7.3 \pm 0.5$ & $12.8 \pm 0.4$ & $6.3$ \\
         $1000$ & $95.8$ & $80.0 \pm 2.4$ & $28.6 \pm 3.3$ & $74 \pm 0.5$ & $7.6 \pm 0.8$ & $12.8 \pm 0.4$ & $9.6$ \\
         \bottomrule
    \end{tabular}
    \end{footnotesize}
    \caption{Performance of \methodsls on benchmark \add{models}, compared across increasing number of contexts. From each context, 2 positive and 2 negative examples were sampled for training. A cutoff of 24 hours was used for each experiment.}
    \label{tab:result_benchmark}
\end{table}
Table~\ref{tab:result_benchmark} reports the performance of \methodsls on benchmark models. The first column represents the number of contexts used to generate the training set. The other columns show the performance of the corresponding learnt model. 
For each experiment we used a cutoff time of 24 hours, however, in the last column we also report the actual time taken to learn the models for which the performances have been reported.
To keep the numbers in perspective, we also report the performance of a random classifier. 
\add{Increasing the number of contexts from $250$ to $500$ and therefore the number of examples, leads to a better model, however increasing it further to $1000$ does not show any significant change across any metric.
With $500$ contexts and a cutoff time of $24$ hours, we achieved $\sim 27\%$ infeasibility, which is a $62\%$ decrease compared to a random classifier. We also observed  around $43\%$ decrease in regret from $12.8\%$ to $7.3\%$. On average it took around $6.3$ hours to achieve this performance, but considering the time taken by experts to model a problem and also the fact that once learnt, these models can be used again and again, we think this is a very reasonable time to learn a model.}

% A major feature of \method is that it does not need to know which negative examples are infeasible and which ones are sub-optimal.  It is natural to ask whether access to this extra information would make learning easier. 

% On the one hand, when the number of context is 100 (3rd column), leading to 400 examples in the training set, \methodmilp fails to learn any model within a cutoff time of $30$ minutes. On the other hand, all implementations of \methodsls are anytime algorithms and always return a learned model. This clearly shows that \methodsls is more scalable than \methodmilp. 

% Among the four implementations of \methodsls, Walk-SAT shows the best performance across all the metrics. It achieves highest score among all SLS implementations, infeasibility does not show much difference when using 50 or 100 contexts, however regret for Walk-SAT clearly shows an improvement over others except when the number of contexts is 25. Seeing these results, going forward we will use just Walk-SAT implementation for further analysis.

\section{Related Work}
\label{sec:related}

Our approach is closely related to constraint learning and acquisition~\cite{bessiere2016new,de2018learning}.  There the goal is to acquire a constraint satisfaction problem (aka constraint network), usually from examples of feasible and infeasible \add{assignments}.  However, the issue of learning from contextual examples, which are pervasive in real-world decision making, is usually ignored.  One exception is QuAcq~\cite{bessiere2013constraint}, which acquires hard constraints from membership queries about partial assignments.  QuAcq, however, is allowed to \emph{choose} informative partial assignments, while in our (harder) case the contexts are fixed by the environment.  In addition, QuAcq does not support acquiring optimization problems and only handles a restricted class of contexts, namely partial assignments.  Context-dependent examples have also been considered in logic-based learning of Answer Set Programs, but the links to combinatorial optimisation and statistical learning are missing~\cite{law2016iterative}.

%Contexts also generalize the part-wise feedback introduced in~\cite{dragone2018decomposition} and used to simplify interaction with a human decision maker;  their results however only work in the online learning case and do not trivially transfer to the problem of learning hard constraints.  Further, in their problem the (analogue of our) contexts is designed and fixed beforehand, not determined by the environment as in our case.   Notice that positive and negative examples are more common in offline learning - which we target.

Most approaches in constraint learning acquire hard constraints only by searching the version space with one-directional~\cite{beldiceanu2012model} or bi-directional search~\cite{bessiere2016new}.  \methodmilp leverages ideas from syntax-guided synthesis~\cite{alur2018search}, where learning is cast as a proper satisfaction or optimization problem and tackled with a solver.  This strategy has been used for learning rules~\cite{malioutov2018mlic}, hybrid logical-numerical formulas~\cite{kolb2018learning}, Bayesian networks~\cite{berg2014learning}, and causal models~\cite{hyttinen2013discovering}.  Stochastic local search heuristics have also received some attention as tools for structure learning of machine learning models, see for instance~\cite{kask1999stochastic,ghallab2008structure}.

Approaches that learn soft constraints build on machine learning techniques such as regression~\cite{rossi2004acquiring}, learning-to-rank~\cite{pawlak2017automatic}, or structured prediction~\cite{teso2017structured}, depending on the kind of available supervision.  However, these methods require the hard constraints to be given.

Other related areas include structured prediction~\cite{mcallester2007generalization,london2016stability} and contextual combinatorial bandits~\cite{li2010contextual}.  In these settings the supervision consists of input-output pairs for which the partitioning of variables into inputs and outputs is fixed.  This schema is much more restrictive than general context-specific examples.  In addition, in structured prediction there is no support for learning hard constraints.

%Our implementation is similar in spirit to INCAL~\cite{kolb2018learning}, a solver-based approach to learning Satisfiability Modulo Linear Real Arithmetic~\cite{barrett2009satisfiability}.  The main differences to INCAL are that the latter (1) is limited to learning hard constraints, albeit with continuous variables and parameters, and (2) it casts learning as a pure satisfaction problem, which cannot be done in our case since in MAX-SAT computing entailment involves solving an optimization problem.  Still, like INCAL, learning is cast as an inference problem and some techniques introduced for INCAL, like incremental learning and parameter-free search could be used to improve the performance of~\method.

%We remark that, although MPE inference in Bayesian networks and Markov Logic Networks can be cast as weighted MAX-SAT~\cite{park2002using,richardson2006markov}, such models are usually learned in a generative manner and -- even when learned discriminatively -- from complete, context-independent examples only.  Structure and parameter learning are also done separately or sequentially.
\add{Maximum a posteriori} inference in Bayesian networks and Markov Logic Networks \add{(MLNs}) can be cast as weighted MAX-SAT~\cite{park2002using,richardson2006markov}. However, MLNs are not learned from contextual examples where both context $x$ and completion $xy$ are provided. Rather the $xy$ are sampled from the underlying probability distribution and it is not assumed that the examples are optimal (i.e.,  maximally likely).
 
One rather recent direction of research are differentiable layers able to implicitly acquire and solve optimization problems within deep networks. Closest to our work, SATNet~\cite{SATNet} can be used in neural nets as a module to learn MAX-SAT models in an end-to-end manner.
%They use block coordinate descent methods to compute the gradients. Doing so they can learn a network that understands the logical structure.
A major difference between our approach and SATNet is that the latter does not really output a proper MAX-SAT model:  the model is implicitly captured by the parameters of the layer and cannot be read off directly.  Another difference is that SATNet layers are also responsible for \emph{solving} the acquired model, and it does so using an approximate \add{semi-definite programming} relaxation.  This heuristic cannot compete in precision or scalability with MAX-SAT solvers.  Another downside of SATNet is that it is restricted to inferring a single, fixed solution, even for problems with multiple ones.  In contrast, \method produces a full-fledged MAX-SAT model that can be easily used to generate multiple solutions using state-of-the-art solvers, \add{moreover experts can even add new contexts as constraints to generate solutions under new temporary restrictions}.  SATNet also learns from complete \add{assignments} only and has no support for contextual examples.

Finally, in terms of theory, this is the first work in the field of constraint learning that provides PAC learnability guarantees.

\section{Conclusion and Future Work}
\label{sec:conclusion}

We introduced the novel learning task of acquiring combinatorial optimization models from contextual data, focusing specifically on MAX-SAT models.  Our analysis shows that ERM provably learns low-regret MAX-SAT models from context-specific examples in both realizable and agnostic setting. It works even in the noisy setting as long as the noise is random and the probability of occurance is less than $50\%$. These results justify our ERM-based implementations, \methodmilp and \methodsls.  The first implementation relies on an approximate but tight MILP encoding to perform learning, while the second one uses stochastic local search. \methodmilp shows good performance when using small number of examples for learning, but fails to scale for large training set. However, increasing the size of the training set shows significant improvement in the quality of the learnt model. \methodsls utilises large training set much more efficiently to learn better models and hence scales much better compared to \methodmilp. Also, if time is of the essence, \methodsls provides the flexibility to learn a model given any cutoff time, however this might decrease the quality of the learnt model.     

There are three clear directions in which this work can be extended.  First, since nothing in our theory relies on the variables being Boolean nor on the constraints being propositional formulas, our results transfer to general combinatorial optimization problems with linear objective functions, e.g., weighted constraint satisfaction problems~\cite{rossi2004acquiring}.  With some effort, it should be possible to generalize them even further to hybrid discrete-continuous optimization frameworks like maximum satisfiability modulo linear arithmetic~\cite{barrett2009satisfiability}, in which the decision variables can be also real valued and the constraints may include linear inequalities.  In practice, this would require, first and foremost, to revise the strategy used by \methodsls for evaluating and exploring the neighborhood of a candidate model.  Second, while our theory does apply to the agnostic and the uniform noisy settings, it would be useful to extend it to deal structured noise, like sub-optimal examples.  One option is to make use of counterfactual risk minimization~\cite{swaminathan2015counterfactual}, an alternative to ERM in which the sampling bias due to irrational annotators is taken into account.  Finally, it is worth lifting the restriction that context annotations are always available, complete, and noiseless.  This could be handled by treating the context as a latent variable and optimizing it to infer their unobserved values.  While very useful, these extensions are highly non-trivial and left to future work.

\subsection*{Acknowledgments}

% For arxiv only
We thank the reviewers for helping to improve the quality of the manuscript.
This work has received funding from the European Research Council (ERC) under the European Union?s Horizon 2020 research and innovation programme (grant agreement No.  [694980] SYNTH: Synthesising Inductive Data Models).  The research of ST was partially supported by TAILOR, a project funded by EU Horizon 2020 research and innovation programme under GA No 952215.

\bibliographystyle{elsarticle-num} 
\bibliography{paper}

\newpage
\appendix

\section{Learnability in a Fixed Context (Section~\ref{sec:intracontext})}

Here we show that MAX-SAT classifiers applied to a specific context $\calH_\context$ have finite VC dimension and are therefore learnable from context-specific examples.  We start by building an easier-to-handle superset $\calG_\context \supseteq \calH_\context$, as follows:
\begin{align*}
    \calG_\context \defeq \{
        & \Ind{(\vx \models \phi(\vc) \land \context) \land f_{\vw}(\vx) \ge \max_{\vx' \,\models\, \phi(\vc') \,\land\, \context} f_{\vw}(\vx')}
    \\
        & : \vw \in [-1,1]^m, \; \vc, \vc' \in \{0, 1\}^m \}
\end{align*}
By construction, $\add{\VC}(\calG_\context) \ge \add{\VC}(\calH_\context)$.  Indeed, fixing $\vc = \vc'$ recovers the definition of $\calH_\context$.  The VC dimension of $\calG_\context$ can be bounded in terms of two simpler sets defined as:
\begin{align*}
    \calG_{\context,\text{hard}} \defeq \{
        & \Ind{ \vx \models \phi(\vc) \land \context } : \vc \in \{0,1\}^m \},
\end{align*}
and:
\begin{align*}
    \calG_{\context,\text{soft}} \defeq \{
        & g_{\vc',\vw}(\cdot, \context) : \vc' \in \{0,1\}^m, \vw \in [-1,1]^m \},
    \\
        & g_{\vc',\vw}(\vx, \context) \defeq \Ind{ f_{\vw}(\vx) \ge \max_{\vx' \models \phi(\vc') \land \context} f_{\vw}(\vx') }
\end{align*}
Our first result relies on a well-known bound on the VC-dimension of intersections of hypothesis classes, reported here for ease of reference:

\begin{theorem}[Theorem~1.1 of~\cite{vandervaart2009note}]
\label{thm:intersectionvc}
Let $\calH_1, \ldots, \calH_l$ be hypothesis classes and $\calH = \bigcap_{j=1}^l \calH_j$ be their intersection.  Then:
$$
    \textstyle
    \add{\VC}(\calH) \le c_1 \log(c_2 l) \sum_{j=1}^l \add{\VC}(\calH_j)
$$
where $c_1 \approx 2.3$ and $c_2 \approx 3.9$.
\end{theorem}

\begin{lemma}
\label{lemma:1}
For any \context and $u < 5$, it holds that:
$$
    \add{\VC}(\calG_\context) \le u (\add{\VC}(\calG_{\context,\text{hard}}) + \add{\VC}(\calG_{\context,\text{soft}}))
$$
\begin{proof}
    Each and every hypothesis in $\calG_\context$ is the conjunction of a concept from $\calG_{\context,\text{hard}}$ with a concept from $\calG_{\context,\text{soft}}$.  The claim follows by applying Theorem~\ref{thm:intersectionvc}.
\end{proof}
\end{lemma}

The remaining step is to prove that $\calG_{\context,\text{hard}}$ and $\calG_{\context,\text{soft}}$ both have bounded VC dimension.
It is clear that $|\calG_{\context,\text{hard}}| \le 2^m$ and therefore $\add{\VC}(\calG_{\context,\text{hard}}) \le m$.  The answer for $\calG_{\context,\text{soft}}$ is given by the following lemma:

\begin{lemma}
For any \context, it holds that $\add{\VC}(\calG_{\context,\text{soft}}) \le m + 1$.
\begin{proof}
$\calG_{\context,\text{soft}}$ can be rewritten as:
\begin{align*}
        & \calG_{\context,\text{soft}} = \{ g_{\vc',\vw}(\cdot, \context) : \vc' \in \{0,1\}^m, \vw \in [-1,1]^m \}
    \\
        & \textstyle = \{ \Ind{ f_{\vw}(\vx) \ge \max_{\vx' \models \phi(\vc') \land \context} f_{\vw}(\vx')} : \vc', \vw \}
    \\
        & \subset \{\Ind{f_{\vw}(\vx) \ge b} : \vw \in [-1,1]^m, b\in\bbR \}
\end{align*}
Here $\vc'$ and $\vw$ implicitly range over $\{0, 1\}^m$ and $[-1, 1]^m$, respectively.  The final set is a subset of $\calL$, the class of linear classifiers over the $m$ features $\Ind{\vx \models \phi_j(\vx)}$. This shows that $\add{\VC}(\calG_\text{soft}) \le \add{\VC}(\calL) \le m + 1$ (Theorem 9.3 of \cite{shalev2014understanding}).
\end{proof}
\end{lemma}

Lemma~\ref{lemma:1} clearly implies Theorem~\ref{thm:ci}. Now combining Theorem~\ref{thm:uc} and Theorem~\ref{thm:ci} shows that:
\ThmUC*

\section{From Classification to Optimization (Section~\ref{sec:clftoopt})}

\ThmRegretBound*
\begin{proof}
First, notice that:
\begin{equation}
    \sum_{\vx : h(\vx,\context) = 1} \Ind{h(\vx, \context) \ne h^*(\vx, \context)}
    \le
    \frac{1}{\minprob} L_{\distr,\context}(h)
    \label{eq:ubrisk}
\end{equation}
Indeed, the risk can be lower bounded as:
\begin{align*}
    L_{\distr,\context}(h)
        & \textstyle = \sum_{\vx \in \{0,1\}^n} \Ind{h(\vx,\context) \ne h^*(\vx,\context)} \distr(\vx\,|\,\context) \nonumber
    \\
        & \textstyle \ge \sum_{\vx \,:\, h(\vx,\context) = 1} \Ind{h^*(\vx,\context) = 0} \distr(\vx\,|\,\context) \nonumber
    \\
        & \textstyle \ge \minprob \sum_{\vx \,:\, h(\vx,\context) = 1} \Ind{h^*(\vx,\context) = 0}
\end{align*}
Second, notice that $\vx$ contributes to the average regret iff $h(\vx, \context) = 1 \land h^*(\vx, \context) = 0$:  if $\vx$ is feasible w.r.t. $\phi(\vc^*)$, it contributes a factor $(f_{\vw^*}(\vx^*) - f_{\vw^*}(\vx))$, else it contributes $\wcr$.  Thus, ignoring constants, the regret can be written as the sum of two terms.  The first one is:
\begin{align}
        & \sum_{\vx \,:\, h(\vx,\context) = 1} \Ind{h^*(\vx,\context) = 0 \land \vx \models \phi(\vc^*)} (f_{\vw^*}(\vx^*) - f_{\vw^*}(\vx)) \nonumber
    \\
        & \le \left[ \max_{\vx} \; f_{\vw^*}(\vx^*) -  f_{\vw^*}(\vx) \right] \sum_{\vx : h(\vx,\context) = 1} \!\!\!\!\!\! \Ind{h^*(\vx,\context) = 0 \land \vx \models \phi(\vc^*)} \nonumber
    % \\
    %     & \le \left[ \max_{\vx} \; \sum_{j = 1}^m w^{*}_j \Ind{\vx^{*} \models \phi_j} -  \sum_{j = 1}^m w^{*}_j \Ind{\vx \models \phi_j} \right] \sum_{\vx : h(\vx,\context) = 1} \!\!\!\!\!\! \Ind{h^*(\vx,\context) = 0 \land \vx \models \phi(\vc^*)} \nonumber
    \\
        & = \left[ \max_{\vx} \; \sum_{j = 1}^m w^{*}_j (\Ind{\vx^{*} \models \phi_j^*} - \Ind{\vx \models \phi_j^*}) \right] \sum_{\vx : h(\vx,\context) = 1} \!\!\!\!\!\! \Ind{h^*(\vx,\context) = 0 \land \vx \models \phi(\vc^*)} \nonumber
    \\
        & \le  \|\vw^*\|_{1}  \sum_{\vx : h(\vx,\context) = 1} \!\!\!\!\!\! \Ind{h^*(\vx,\context) = 0 \land \vx \models \phi(\vc^*)}
        \label{eq:annswhackyadventures}
        % & \le \left[ \max_{\vx,\vx'} \; f_{\vw^*}(\vx) -  f_{\vw^*}(\vx') \right] \sum_{\vx : h(\vx,\context) = 1} \!\!\!\!\!\! \Ind{h^*(\vx,\context) = 0 \land \vx \models \phi(\vc^*)} \nonumber
\end{align}
where the last step follows from H\"older's inequality.  The second one is:
\begin{equation}
    \wcr \sum_{\vx \,:\, h(\vx,\context) = 1} \Ind{h^*(\vx,\context) = 0 \land \vx \not\models \phi(\vc^*)} \label{eq:bobswackyadventures}
\end{equation}
The summations in Eq.~\ref{eq:annswhackyadventures} and Eq.~\ref{eq:bobswackyadventures} are upper bounded by the left hand side of Eq.~\ref{eq:ubrisk}.  Summing the two quantities and dividing by $|\Optima(\context)|$ gives the average regret and proves the claim.
\end{proof}

\section{Generalizing across Contexts (Section~\ref{sec:intercontext})}

\mixtotop*
\begin{proof}
By definition of $\errormatch$ and $\#$, it holds that:
\begin{align*}
    \Ind{h(\vx, \context_\text{t}) \ne h^*(\vx,\context_\text{t})}
    & = \frac{1}{\#(\context_\text{t}, \vx)} \sum_{\context \in \Contexts \,:\, \errormatch(\context_\text{t}, \vx, \context)} \Ind{h(\vx, \context) \ne h^*(\vx, \context)}
\end{align*}
where $\#(\context_\text{t}, \vx) > 0$ because \Contexts is representative.  Plugging this into the definition of risk gives:
\begin{align*}
    L_{\distr, \context_\text{t}}(h) & = \sum_{\vx \in \{0, 1\}^n} \Ind{h(\vx, \context_\text{t}) \ne h^*(\vx,\context_\text{t})} \distr(\vx \,|\, \context_\text{t})
    \\
    & = \sum_{\vx} \left( \sum_{\context \in \Contexts \,:\, \errormatch(\context_\text{t}, \vx, \context)} \frac{\Ind{h(\vx, \context) \ne h^*(\vx, \context)}}{\#(\context_\text{t}, \vx)} \right) \distr(\vx \,|\, \context_\text{t})
    \\
    & = \sum_{\context \in \Contexts} \sum_{\vx \,:\, \errormatch(\context_\text{t}, \vx, \context)} \Ind{h(\vx, \context) \ne h^*(\vx,\context)} \frac{\distr(\vx \,|\, \context_\text{t})}{\#(\context_\text{t}, \vx)}
\end{align*}
Letting:
$$
    \beta_\context = \max_{\vx \models \context} \frac{1}{\#(\context_\text{t}, \vx)} \frac{\distr(\vx \,|\, \context_\text{t})}{\distr(\vx \,|\, \context)}
$$
we can write:
$$
    L_{\distr,\context_\text{t}}(h) \le \sum_{\context \in \Contexts} \beta_\context \sum_{\vx \,:\, \errormatch(\context_\text{t}, \vx, \context)} \Ind{h(\vx, \context) \ne h^*(\vx,\context)} \distr(\vx \,|\, \context) = \sum_{\context \in \Contexts} \beta_\context L_{\distr,\context}(h)
$$
For each $\beta_\context$, the max runs over those $\vx$'s that satisfy \context, hence $\distr(\vx | \context)$ is always non-zero, implying that each $\beta_\context$ is finite.
\end{proof}

\lastone*
\begin{proof}
%
% $$
%     \textstyle
%     \sum_{(\vx,\context) \in \dataset} \Ind{h(\vx, \context) \ne h^*(\vx, \context)} = 0
% $$
%
Corollary~\ref{thm:uc} guarantees that for every $\epsilon_\context$, $\delta_\context \in (0, 1)$, there exists an integer $s_\context(\epsilon_\context,\delta_\context)$ such that if \dataset includes at least $s_\context(\epsilon_\context,\delta_\context)$ examples for context \context, then:
\begin{align}
    \prob(L_{\distr,\context}(h) > L_{\dataset,\context}(h) + \epsilon_\context) \le \delta_\context && \context \in \Contexts
    \label{eq:dunno}
\end{align}
Let $l$ represent the number of mistakes made by the empirical risk minimizer on \dataset, that is, $l=\min_{h \in \calH}\sum_{\context \in \Contexts}L_{\dataset,\context}(h)$.
The second part of the proof involves upper bounding the probability that $h$ has large risk in the target context $\context_\text{t}$. Let $\beta=\max_{\context \in \Contexts}\beta_{\context}$, then:
\begin{align}
    \textstyle \prob(L_{\distr,\context_\text{t}}(h) > l+\epsilon)
    & \textstyle \le \prob(\beta \sum_{\context \in \Contexts} L_{\distr,\context}(h) > l+\epsilon) \nonumber
    \\
    & \textstyle \le \prob(\beta |\Contexts| \max_{\context \in \Contexts} L_{\distr,\context}(h) > l+\epsilon) \nonumber
    \\
    & \textstyle = \prob(\max_{\context \in \Contexts} L_{\distr,\context}(h) > \frac{l+\epsilon}{\beta |\Contexts|}) \nonumber
    \\
    & \textstyle \le \prob(\bigvee_{\context \in \Contexts} ( L_{\distr,\context}(h) > \frac{l+\epsilon}{\beta |\Contexts|} ) ) \nonumber
    \\
    & \textstyle \le \sum_{\context \in \Contexts} \prob( L_{\distr,\context}(h) > \frac{l+\epsilon}{\beta |\Contexts|})
    \label{eq:sumofrisks}
\end{align}
The first step follows from Lemma~\ref{thm:mixtotop}. By Eq.~\ref{eq:dunno}, the last expression can be made smaller than any $\delta$ by adding enough context-specific examples to \dataset.   In particular, having at least:
$$
    t_\context(\epsilon,\delta) = s_\context\left(\frac{l+\epsilon}{\beta |\Contexts|}, \delta_\context\right)
$$
examples for every context guarantees that Eq.~\ref{eq:sumofrisks} is less than any $\delta = \sum_{\context \in \Contexts} \delta_\context$.  This concludes the proof.
\end{proof}

\section{MILP Encoding}
\label{appendix:milp}

The variables used in the MILP encoding are explained in Table~\ref{tab:vars} and the MILP encoding itself if given in Figures~\ref{fig:milp1} and~\ref{fig:milp2}.
Note, that the MILP encoding makes the following assumptions: 1) contexts are partial assignments, i.e., conjunctions of literals; 2) the MAX-SAT problem is in CNF format, i.e., both hard and soft constraints are clauses (disjunctions of literals); and 3) weights are in the interval $[0, 1]$.

\newcommand{\ccov}{\mathrm{ccov}}
\newcommand{\cov}{\mathrm{cov}}
\newcommand{\opt}{\mathrm{opt}}
\newcommand{\covp}{\mathrm{cov}'}
\newcommand{\sel}{a}
\newcommand{\lit}{\ell}
\newcommand{\lits}{L}
\newcommand{\xp}{\vx'}
\newcommand{\wz}{\mathrm{wz}}
\newcommand{\eps}{\epsilon}

\begin{table*}
\footnotesize
	\caption{\label{tab:vars} Names, types and explanations of the variables used in the MILP encoding.}
	\begin{tabular}{rccp{8cm}}
	    \toprule
		&\textbf{Symbol} & \textbf{Type} & \textbf{Description}\\
		\midrule
		\emph{Constants}
		&$M$ & int & Large constant to disable constraints $M = 2\cdot(m + 1)$\\
		&$\eps$ & real & Small constant to model strict inequality $(\eps = 10^{-2})$\\
		&$m$ & int & Number of constraints\\
		&$s$ & int & Number of examples\\
		&$n$ & int & Number of Boolean variables\\
		&$L$ & set & All literals $(\{X_i | i =1..n\} \cup \{\lnot X_i | i=1..n\})$\\
		&$L^+$ & set & All positive literals $(\{X_i | i =1..n\}$\\
		&$L^-$ & set & All negative literals $(\{\lnot X_i | i =1..n\}$\\
		&$\Contexts$ & set & All unique contexts occuring in the examples\\
		&$\context$ & set & A context (a set of literals)\\
		&$|\context|$ & int & Number of literals in context~\context\\
		&$\context^+$ & set & All positive literals in the context~\context\\
		&$\context^-$ & set & All negative literals in the context~\context\\\hline
		\emph{Variables}
		&$\ccov^\context$ & bool & True iff context~\context feasible\\
		&$\xp^\context$ & \add{assignment} & The optimal \add{assignment} in context~\context \\
		&$\xp^\context_\lit$ & bool & True iff literal~$\lit$ is true in \add{assignment}~$\xp^\context$\\
		&$\cov^\context_{j\lit}$ & bool & True iff clause~$j$ contains literal~$\lit$ and $\xp^\context_\lit$ is true\\
		&$\sel_{j\lit}$ & bool & True iff clause~$j$ contains literal~$\lit$\\
		&$\cov^\context_{j}$ & bool & True iff clause~$j$ covers $\xp^\context$, i.e., $\xp^\context$ satisfies clause~$j$\\
		&$c_{j}$ & bool & True if clause~$j$ is a hard constraint, otherwise it is a soft constraint \\
		&$\covp^\context_{j}$ & bool & True iff clause~$j$ is a soft constraint or clause~$j$ covers $\xp^\context$ $(\lnot c_j \lor \cov^\context_{j})$ \\
		&$\cov^\context$ & bool & True iff $\xp^\context$ satisfies all hard constraints and the context is feasible $(\ccov^\context \land \bigwedge_{j=1}^m \covp^\context_j)$\\
		&$w_j$ & real $\in [0, 1] $ & The weight assigned to clause~$j$\\
		&$\wz_j$ & bool & True iff the weight assigned to clause~$j$ is $0$ $(\Ind{w_j = 0})$\\
		&$w^\context_j$ & real $\in [0, 1]$ & $w_j$ if the clause~$j$ is soft and covers $\xp^\context$, $0$ otherwise $(w^\context_j = w_j \cdot \Ind{\lnot c_j} \cdot \Ind{\cov^\context_j})$ \\
		&$\gamma_\context$ & real & The optimal value of any point~($\xp^\context$) in context~$\context$\\
		&$w_{jk}$ & real $\in [0, 1]$ & $w_j$ if clause~$j$ is soft and covers $\vx_k$, $0$ otherwise $(w^\context_j = w_j \cdot \Ind{\lnot c_j} \cdot \Ind{\cov_{jk}})$\\
		&$\cov_{jk}$ & bool & True iff clause~$j$ covers example~$k$\\
		&$\opt_k$ & bool & True iff the $k$th example ($\vx_k$) is optimal\\
		&$\covp_{jk}$ & bool & True iff clause~$j$ is a soft constraints or clause~$j$ covers example~$k$ $(\lnot c_j \lor \cov_{jk})$\\
		&$\vx_{k\lit}$ & bool & True iff literal~$\lit$ is true in example~$k$\\
		&$\cov_{k}$ & bool & True iff $\vx_k$ satisfies all hard constraints $(\bigwedge_{j=1}^m \covp_{jk})$\\
		\bottomrule
	\end{tabular}
\end{table*}

\begin{figure*}
\footnotesize
\begin{align}
	\mathrm{max} \;\; &\sum_{\context \in \Contexts} \gamma_\context &\\
    \mathrm{s.t.} \;\;&\ccov^{\context} \leq \xp^{\context}_{\lit} &\forall \context \in \Contexts, \lit \in \context^+\\
    &\ccov^{\context} \leq (1 - \xp^{\context}_{\lit}) &\forall \context \in \Contexts, \lit \in \context^-\\
    &\ccov^{\context} \geq \sum_{\lit \in \context^+} \xp^{\context}_\lit + \sum_{\lit \in \context^-} (1 - \xp^{\context}_l) - |\context| + 1 &\forall \context \in \Contexts\\
    &\cov^\context_{j\lit} \leq \sel_{j\lit} &\forall \context \in \Contexts, j = 1..m, \lit \in \lits\\
    &\cov^\context_{j\lit} \leq \xp^{\context}_\lit &\forall \context \in \Contexts, j = 1..m, \lit \in \lits^+\\
    &\cov^\context_{j\lit} \geq \sel_{j\lit} + \xp^{\context}_\lit - 1 &\forall \context \in \Contexts, j = 1..m, \lit \in \lits^+\\
    &\cov^\context_{j\lit} \leq 1 - \xp^{\context}_\lit &\forall \context \in \Contexts, j = 1..m, \lit \in \lits^-\\
    &\cov^\context_{j\lit} \geq \sel_{j\lit} + (1 - \xp^{\context}_\lit) + 1 &\forall \context \in \Contexts, j = 1..m, \lit \in \lits^-\\
    &\cov^\context_j \geq \cov^\context_{j\lit} &\forall \context \in \Contexts, j = 1..m, \lit \in \lits\\
    &\cov^\context_j \leq \sum_{\lit \in \lits} \cov^\context_{j\lit} &\forall \context \in \Contexts, j = 1..m\\
	&\covp^\context_j \geq \cov^\context_j &\forall \context \in \Contexts, j = 1..m\\
	&\covp^\context_j \geq 1 - c_j &\forall \context \in \Contexts, j = 1..m\\
	&\covp^\context_j \leq \cov^\context_j + (1 - c_j) &\forall \context \in \Contexts, j = 1..m\\
	&\cov^\context \leq \ccov^\context &\forall \context \in \Contexts\\
	&\cov^\context \leq \covp^\context_j &\forall \context \in \Contexts, j = 1..m\\
	&\cov^\context \geq \sum_{j = 1}^m \covp^\context_j + \ccov^\context - m &\forall \context \in \Contexts\\
	&w^\context_j \leq \cov^\context_j &\forall \context \in \Contexts, j = 1..m\\
	&w^\context_j \leq (1 - c_j) &\forall \context \in \Contexts, j = 1..m\\
	&w^\context_j \leq w_j + (1 - \cov^\context_j) + c_j &\forall \context \in \Contexts, j = 1..m\\
	&w^\context_j \geq w_j - (1 - \cov^\context_j) - c_j &\forall \context \in \Contexts, j = 1..m\\
	&\gamma_\context \leq \sum_{j = 1}^m w^\context_j &\forall \context \in \Contexts\\
	&\gamma_\context \leq \cov^\context \cdot M &\forall \context \in \Contexts\\
\end{align}
\caption{\label{fig:milp1} Objective function and first part of the MILP encoding (Eq.~10 in the paper)}
\end{figure*}
\begin{figure*}
\footnotesize
\begin{align}
	\mathrm{s.t.} \;\;&w_j \leq (1 - \wz_j) &\forall j = 1..m\\
	&w_j \geq 3 \cdot \eps - \wz_j &\forall j = 1..m\\
	&w_{jk} \leq \cov_{jk} &\forall j = 1..m, k=1..s\\
	&w_{jk} \leq (1 - c_j) &\forall j = 1..m, k=1..s\\
	&w_{jk} \leq w_j + (1 - \cov_{jk}) + c_j &\forall j = 1..m, k=1..s\\
	&w_{jk} \geq w_j - (1 - \cov_{jk}) - c_j &\forall j = 1..m, k=1..s\\
	&\gamma_{\context_k} \leq (\sum_{j=1}^m w_{jk}) + M \cdot (1 - \opt_k) &\forall k=1..s\\
	&\gamma_{\context_k} \geq (\sum_{j=1}^m w_{jk}) + \eps - M \cdot \opt_k &\forall k=1..s\\
	&\covp_{jk} \geq \cov_{jk} &\forall j = 1..m, k=1..s\\
	&\covp_{jk} \geq 1 - c_j &\forall j = 1..m, k=1..s\\
	&\covp_{jk} \leq \cov_{jk} + (1 - c_j) &\forall j = 1..m, k=1..s\\
	&\cov_{jk} \geq \sel_{j\lit} \cdot \vx_{k\lit} &\forall j = 1..m, k=1..s, \lit \in \lits\\
	&\cov_{jk} \leq \sum_{\lit \in \lits} \sel_{j\lit} \cdot \vx_{k\lit} &\forall j = 1..m, k=1..s\\
	&\sel_{j\lit} + \sel_{j\lnot\lit} \leq 1 &\forall j=1..m, \lit \in \lits^+\\
	&\cov_k \leq \covp_{jk} &\forall j=1..m, k=1..s\\
	&\cov_k \geq (\sum_{j=1}^m \covp_{jk}) - (m - 1) &\forall k=1..s\\
	&\cov_k \leq \cov^{\context_k} &\forall k = 1..s\\
	&\cov_k + \opt_k + \cov^{\context_k} = 3 &\forall k \in \{k | k=1..s \land y_k\}\\
	&\cov_k + \opt_k + \cov^{\context_k} \leq 2 &\forall k \in \{k | k=1..s \land \lnot y_k\}
\end{align}
\caption{\label{fig:milp2} Second part of the MILP encoding (Eq.~11 in the paper)}
\end{figure*}

\subsection{Mismatch between original and MILP encodings}
\begin{example}
	Consider the following target model over three variables~$X_1, X_2$ and~$X_3$:
	\begin{align}
		& X_1 \lor \lnot X_3\\
		\mathit{1.0}\;\; & \lnot X_1
	\end{align}
	and the following examples to learn from (literal assignments fixed by the context are in bold):
	\begin{center}
		\begin{tabular}{ccc|c}
			$X_1$ & $X_2$ & $X_3$ & $y$\\\hline\hline
			$1$ & $1$ & $1$ & $0$\\
			$0$ & $1$ & $0$ & $1$\\
			$1$ & $1$ & $\mathbf{1}$ & $1$\\
			$1$ & $0$ & $\mathbf{1}$ & $1$\\
			$0$ & $\mathbf{1}$ & $\mathbf{0}$ & $1$\\
			$1$ & $\mathbf{1}$ & $\mathbf{0}$ & $0$\\
			$0$ & $\mathbf{0}$ & $\mathbf{0}$ & $0$\\
		\end{tabular}
	\end{center}
	In this case, the MILP encoding will learn the following model:
	\begin{align}
		\mathit{1.0}\;\; & \lnot X_1\\
		\mathit{1.0}\;\; & \lnot X_1 \lor X_3
	\end{align}
	which mislabels the positive examples~$[1, 1, \mathbf{1}]$ and~$[1, 0, \mathbf{1}]$.
	We can examine the gamma values for the true model~($\gamma^T$), the actual values for the learned model~($\gamma^{L}$) as well as those reported by the MILP solution~($\hat\gamma^L$):
	\begin{center}
		\begin{tabular}{c|cccc|c}
			$\gamma \backslash \context$ & $\context_\text{t}$ & $X_3$ & $X_2 \land \lnot X_3$ & $\lnot X_2 \land \lnot X_3$ & $\sum_\context \gamma_\context$\\\hline\hline
			$\gamma^T_\context$ & $1$ & $0$ & $1$ & $1$ & $3$\\
			$\gamma^L_\context$ & $2$ & $2$ & $2$ & $2$ & $8$\\
			$\hat\gamma^L_\context$ & $2$ & $1$ & $2$ & $2$ & $7$
		\end{tabular}
	\end{center}
	Setting gamma correctly (for the learned model) shows that the positive examples~$[1, 1, \mathbf{1}]$ and~$[1, 0, \mathbf{1}]$ are not labeled correctly, as there exists an example ($[0, 1, \mathbf{1}]$) in the same context that is feasible and obtains a higher value $f_{\vw}([0, 1, \mathbf{1}]) = 2$.
	Therefore: $$\gamma^L_{X_3} = \mathrm{max}_{\vx \models \phi \land X_3} f_{\vw}(\vx) = 2 > f_{\vw}([1, 1, \mathbf{1}]) = 1$$

	However, by setting~$\hat\gamma^L_{X_3}$ sub-optimally, those examples are labeled as positives in the MILP encoding: $$\hat\gamma^L_{X_3} \leq f_{\vw}([1, 1, \mathbf{1}]) = f_{\vw}([1, 0, \mathbf{1}]) = 1$$ and the model can obtain a larger value for $\sum_\context \gamma_\context$ than possible when using, for example, the target model.
\end{example}

\section{Generating neighbours for \methodsls}
\label{appendix:neighbours}
\methodsls learns MAX-SAT models, which means the set of constraints $\Phi$ is a set of disjunctions of literals. For any constraint $\phi_i$ and literal $l$, we use $\phi_i - l$ to represent the disjunction after removal of $l$, if present. For example, if $\phi_i=x_1 \lor \neg x_2 \lor x_3$ then $\phi_i - (\neg x_2)=x_1 \lor x_3$. Similarly $\phi_i + l$ represents the disjunction after adding the literal. So, $\phi_i + x_4 = x_1 \lor \neg x_2 \lor x_3 \lor x_4$. To generate the set of neighbours for any model $\program$, first we randomly pick an example $\vx$, which is incorrectly classified by $\program$. The set of neighbours is then generated based on the true and predicted label of $\vx$. We will use $\vx^*$ to represent an optimal example for $\program$.
\begin{itemize}
    \item If true label is positive while the predicted label is infeasible:
    \begin{enumerate}
        \item Make changes in any hard constraint $\phi_i$ such that $\vx \not \models \phi_i$
    \end{enumerate}
\end{itemize}
\begin{itemize}
    \item If true label is positive while the predicted label is sub-optimal:
    \begin{enumerate}
        \item Pick any hard constraint $\phi_i$ such that $\vx \models \phi_i$. Pick a literal $l$ such that $\vx^* \models l$ and replace $\phi_i$ by $\phi_i - l$
        
        \item Pick any soft constraint $\phi_i$ such that $\vx \models \phi_i$ and $\vx^* \not \models \phi_i$. Either make it a hard constraint or increase the weight $\vw_i$ to $(\vw_i + 1) / 2$
        
        \item Pick any soft constraint $\phi_i$ such that $\vx \not \models \phi_i$ and $\vx^* \models \phi_i$. Then do one of these three: 1) Pick a literal $l$ such that $\vx \models l$ and replace $\phi_i$ by $\phi_i + l$, 2) Pick a literal $l$ such that $\vx^* \models l$ and replace $\phi_i$ by $\phi_i - l$ or 3) Reduce the weight $\vw_i$ to $\vw_i / 2$
        
        \item Pick any soft constraint $\phi_i$ such that $\vx \not \models \phi_i$ and $\vx^* \not \models \phi_i$. Pick a literal $l$ such that $\vx \models l$ but $\vx^* \not \models l$, then replace $\phi_i$ by $\phi_i + l$
        
        \item Pick any soft constraint $\phi_i$ such that $\vx \models \phi_i$ and $\vx^* \models \phi_i$. Pick a literal $l$ such that $\vx \not \models l$ but $\vx^* \models l$, then replace $\phi_i$ by $\phi_i + l$
    \end{enumerate}
\end{itemize}
\begin{itemize}
    \item If true label is negative while the predicted label is positive:
    \begin{enumerate}
        \item Pick any hard constraint $\phi_i$ such that $\vx \models \phi_i$. Pick a literal $l$ such that $\vx \models l$ and replace $\phi_i$ by $\phi_i - l$
        
        \item Pick any soft constraint $\phi_i$ such that $\vx \not \models \phi_i$ and $\vx^* \not \models \phi_i$. Either make it a hard constraint or increase the weight $\vw_i$ to $(\vw_i + 1) / 2$
        
        \item Pick any soft constraint $\phi_i$ such that $\vx \models \phi_i$ and $\vx^* \models \phi_i$. Pick a literal $l$ such that $\vx \models l$, then replace $\phi_i$ by $\phi_i - l$ or reduce the weight $\vw_i$ to $\vw_i / 2$
    \end{enumerate}
\end{itemize}
\end{document}